\documentclass[10pt,twocolumn]{article}

\usepackage{times}
\usepackage{epsfig}
\usepackage{graphicx}
\usepackage{amsmath}
\usepackage{amssymb}
\usepackage{xcolor}
\usepackage{subfig}
\usepackage{url}
\usepackage[normalem]{ulem}
\usepackage{wrapfig}
\usepackage{verbatim}

\renewcommand{\k}{k}

\newcommand{\slack}{\boldsymbol{\xi}}
\newcommand{\x}{\boldsymbol{x}}

\newcommand{\ind}{l}
\newcommand{\betaj}{\beta}
\newcommand{\qed}{{\hfill \large $\Box$}}
{\trivlist\item[\hskip\labelsep{\it #1.}]}
{\hspace*{\fill}\qed\endtrivlist}
\renewcommand{\vec}[1]{\ensuremath{\boldsymbol{#1}}}

\newcommand{\smallparagraph}[1]{{\smallskip\noindent{\textbf{#1}}~~}}

\newcommand{\T}{{\top}}
\newcommand{\vbeta}{{\boldsymbol \beta}}

\newcommand{\valpha}{{\boldsymbol \alpha}}

\newlength{\widthtmp}

\begin{document}
\title{Insights from Classifying Visual Concepts with Multiple Kernel Learning}
\author{Alexander~Binder\footnote{corresponding author, alexander.binder@tu-berlin.de}  \footnote{A. Binder and W. Samek are with Technische Universit\"at Berlin and Fraunhofer FIRST,Berlin, Germany}, Shinichi~Nakajima\footnote{S. Nakajima is with Optical Research Laboratory, Nikon Corporation Tokyo}, Marius~Kloft\footnote{M. Kloft and C. M\"{u}ller are with Technische Universit\"at Berlin, Germany}, Christina~M\"{u}ller, Wojciech~Samek,\\Ulf~Brefeld\footnote{U. Brefeld is with the Universit{\"a}t Bonn, Germany}, Klaus-Robert~M\"{u}ller\footnote{K.-R. M\"uller is with Technische Universit\"at Berlin, Germany, and the Institute of Pure and Applied Mathematics at UCLA, Los Angeles, USA}, and~Motoaki~Kawanabe\footnote{M. Kawanabe is with ATR Research, Kyoto, Japan}}

  %\IEEEcompsocthanksitem A. Binder, M. Kloft, C. M\"{u}ller and W. Samek are with Technische Universit\"at Berlin, Germany; E-mail:
   %      \{alexander.binder@, kloft@, mueller@, wojciech.samek@campus.\}tu-berlin.de.
  %\IEEEcompsocthanksitem K.-R. M\"uller is with Technische Universit\"at Berlin, Germany, and the Institute of Pure and Applied Mathematics at UCLA, Los Angeles, USA; E-mail: klaus-robert.mueller@tu-berlin.de
  %\IEEEcompsocthanksitem S. Nakajima is with Optical Research Laboratory, Nikon Corporation Tokyo, Japan; E-Mail: nakajima.s@nikon.co.jp.
  %\IEEEcompsocthanksitem U. Brefeld is with the Universit{\"a}t Bonn, Germany; E-mail: brefeld@uni-bonn.de.
  %\IEEEcompsocthanksitem M. Kawanabe is with ATR Research, Kyoto, Japan; E-mail: kawanabe@atr.jp.

\maketitle

\begin{abstract}
  Combining information from various image features has become a standard technique in concept recognition tasks. However, the optimal way of fusing the resulting kernel functions is usually unknown in practical applications. Multiple kernel learning (MKL) techniques allow to determine an optimal linear combination of such similarity matrices. Classical approaches to MKL promote sparse mixtures. Unfortunately, so-called 1-norm MKL variants are often observed to be outperformed by an unweighted sum kernel. The contribution of this paper is twofold: We apply a recently developed non-sparse MKL variant to state-of-the-art concept recognition tasks within computer vision. We provide insights on benefits and limits of non-sparse MKL and compare it against its direct competitors, the sum kernel SVM and the sparse MKL. We report empirical results for the PASCAL VOC 2009 Classification and ImageCLEF2010 Photo Annotation challenge data sets. About to be submitted to PLoS ONE.
\end{abstract}

%%%%%%%%%%%%%%%%%%%%%%%%%%%%%%%%%%%%%%%%%%%%%%%%%%%%%%%%%%%%%%%%%%%%%%%%%%%%%%%%%%%%%%%%%%

%\section*{To Do}
%\begin{itemize}
%  \item 
%\end{itemize}

%%%%%%%%%%%%%%%%%%%%%%%%%%%%%%%%%%%%%%%%%%%%%%%%%%%%%%%%%%%%%%%%%%%%%%%%%%%%%%%%%%%%%%%%%%
\section{Introduction}\label{intro}
%%%%%%%%%%%%%%%%%%%%%%%%%%%%%%%%%%%%%%%%%%%%%%%%%%%%%%%%%%%%%%%%%%%%%%%%%%%%%%%%%%%%%%%%%%

A common strategy in visual object recognition tasks is to combine different image representations to capture relevant traits of an image. Prominent representations are for instance built from color, texture, and shape information and used to accurately locate and classify the objects of interest. The importance of such image features changes across the tasks. For example, color information increases the detection rates of stop signs  in images substantially but it is almost useless for finding cars. This is because stop sign are usually red in most countries but cars in principle can have any color. As additional but nonessential features not only slow down the computation time but may even harm predictive performance, it is necessary to combine only relevant features for state-of-the-art object recognition systems. 

We will approach visual object classification from a machine learning perspective. In the last decades, support vector machines (SVM) \cite{Vapnik95,CorVap95,Vapnik98} have been successfully applied to many practical problems in various fields including computer vision \cite{Chapelle99}. Support vector machines exploit similarities of the data, arising from some (possibly nonlinear) measure. The matrix of pairwise similarities, also known as kernel matrix, allows to abstract the data from the learning algorithm \cite{MueMikRaeTsuSch01,book:Schoelkopf+Smola:2002}.  

That is, given a task at hand, the practitioner needs to find an appropriate similarity measure and to plug the resulting kernel into an appropriate learning algorithm. But what if this similarity measure is difficult to find?  We note that \cite{Jaakkola98} and \cite{Zien00} were the first to exploit prior and domain knowledge for the kernel construction. 

In object recognition, translating information from various image descriptors into several kernels has now become a standard technique. Consequently, the choice of finding the right kernel changes to finding an appropriate way of fusing the kernel information; however, finding the right combination for a particular application is so far often a matter of a judicious choice (or trial and error).

In the absence of principled approaches, practitioners frequently resort to heuristics such as uniform mixtures of normalized kernels \cite{Lazebnik06,Zhang07} that have proven to work well. Nevertheless, this may lead to sub-optimal kernel mixtures.

An alternative approach is multiple kernel learning (MKL) that has been applied to object classification tasks involving various image descriptors \cite{sminchisescu_kumar07,GehlerN09}. Multiple kernel learning \cite{Lanckriet04,Bach04,SonRaeSchSch06,RakBacCanGra08} generalizes the support vector machine framework and aims at learning the optimal kernel mixture and the model parameters of the SVM simultaneously. To obtain a well-defined optimization problem, many MKL approaches promote sparse mixtures by incorporating a $1$-norm constraint on the mixing coefficients. Compared to heuristic approaches, MKL has the appealing property of learning a kernel combination (wrt.~the $\ell_1$-norm constraint) and converges quickly as it can be wrapped around a regular support vector machine \cite{SonRaeSchSch06}. However, some evidence shows that sparse kernel mixtures are often outperformed by an unweighted-sum kernel \cite{WSNips}. As a remedy, \cite{KloBreLasSon08,CorMohRos09b} propose $\ell_2$-norm regularized MKL variants, which promote non-sparse kernel mixtures and subsequently have been extended to $\ell_p$-norms \cite{KloBreSonZieLasMue09,KloBreSonZie11}. 

Multiple Kernel approaches have been applied to various computer vision problems outside our scope such multi-class problems \cite{DBLP:conf/cvpr/OrabonaJC10} which require mutually exclusive labels and object detection \cite{vedgulvarzis09,DBLP:conf/cvpr/GalleguillosMBL10} in the sense of finding object regions in an image. The latter reaches its limits when image concepts cannot be represented by an object region anymore such as the \emph{Outdoor},\emph{Overall Quality} or \emph{Boring} concepts in the ImageCLEF2010 dataset which we will use.

In this contribution, we study the benefits of sparse and non-sparse MKL in object recognition tasks. We report on empirical results on image data sets from the PASCAL visual object classes (VOC) 2009 \cite{pascal-voc-2009} and ImageCLEF2010 PhotoAnnotation \cite{imageclef2010} challenges, showing that non-sparse MKL significantly outperforms the uniform mixture and $\ell_1$-norm MKL. Furthermore we discuss the reasons for performance gains and performance limitations obtained by MKL based on additional experiments using real world and synthetic data. 

The family of MKL algorithms is not restricted to SVM-based ones. Another competitor, for example, is Multiple Kernel Learning based on Kernel Discriminant Analysis (KDA) \cite{Yan:icdm09,Yan:cvpr2010}. The difference between MKL-SVM and MKL-KDA lies in the underlying single kernel optimization criterion while the regularization over kernel weights is the same. 

Outside the MKL family, however, within our problem scope of image classification and ranking lies, for example, \cite{DBLP:conf/iccv/CaoLLH09} which uses a logistic regression as base criterion and results in a number of optimization parameters equal to the number of samples times the number of input features. Since the approach in \cite{DBLP:conf/iccv/CaoLLH09} uses a priori much more optimization variables, it poses a more challenging and potentially more time consuming optimization problem which limits the number of applicable features and can be evaluated for our medium scaled datasets in detail in the future. 

Alternatives use more general combinations of kernels such as products with kernel widths as weighting parameters \cite{gehler_kernelfigureitout_cvpr2009,varmababu}. As \cite{varmababu} point out the corresponding optimization problems are no longer convex. Consequently they may find suboptimal solutions and it is more difficult to assess using such methods how much gain can be achieved via learning of kernel weights.

This paper is organized as follows. In Section \ref{MLT}, we briefly review the machine learning techniques used here; The following section\ref{experiment} we present our experimental results on the VOC2009 and ImageCLEF2010 datasets; in Section~\ref{section:disc_toy} we discuss promoting and limiting factors of MKL and the sum-kernel SVM in three learning scenarios.

%%%%%%%%%%%%%%%%%%%%%%%%%%%%%%%%%%%%%%%%%%%%%%%%%%%%%%%%%%%%%%%%%%%%%%%%%%%%%%%%%%%%%%%%%%
\section{Methods}\label{MLT}
%%%%%%%%%%%%%%%%%%%%%%%%%%%%%%%%%%%%%%%%%%%%%%%%%%%%%%%%%%%%%%%%%%%%%%%%%%%%%%%%%%%%%%%%%%

This section briefly introduces multiple kernel learning (MKL), and kernel target alignment. For more details we refer to the supplement and the cited works in it. 

%-------------------------------
\subsection{Multiple Kernel Learning}\label{mkl1}
%-------------------------------

Given a finite number of different kernels each of which implies the existence of a feature mapping $\psi_j:\mathcal{X}\rightarrow\mathcal{H}_j$ onto a hilbert space \[\k_j(\x,\bar\x)=\langle\psi_j(\x),\psi_j(\bar\x)\rangle_{\mathcal{H}_j}\] the goal of multiple kernel learning is to learn SVM parameters $(\vec \alpha,b)$ and linear kernel weights $K=\sum_l \beta_l k_l$ simultaneously.

This can be cast as the following optimization problem which extends support vector machines \cite{CorVap95,book:Schoelkopf+Smola:2002}
\begin{align}\label{zien}
   \min_{\vbeta,\vec v, b, \vec\slack}  \quad & \frac{1}{2}\sum_{j=1}^m \frac{\vec 
v_j'\vec v_j}{\beta_j} + C\Vert\slack\Vert_1 \\
   \text{s.t.} \quad & \forall i: ~ y_i\left( \sum_{j=1}^m \vec v_j'\psi_j
(\vec x_i) +b \right)\geq 1-\xi_i \nonumber\\
   & \slack\geq \vec 0 ~ ; \quad \vbeta\geq \vec 0 ~ ; \quad \Vert\vbeta\Vert_p\leq 1.\nonumber
\end{align}
The usage of kernels is permitted through its partially dualized form:
\begin{align*}%\label{minmax}
  \min_{\vbeta} \, \max_{\valpha} \quad & \sum_{i=1}^n\alpha_i -\frac{1}{2}\sum_{i,\ind=1}^n\alpha_i\alpha_\ind y_iy_\ind\sum_{j=1}^m\betaj_j\k_j(\vec x_i,\vec x_\ind) \\
  \text{s.t.} \quad & \forall_{i=1}^n:\,\, 0\leq\alpha_i\leq C; \quad \sum_{i=1}^n y_i\alpha_i=0; \\
&\forall_{j=1}^m: \,\betaj_j\geq 0; \quad \Vert\vbeta\Vert_p \leq 1 \nonumber.
\end{align*}
For details on the solution of this optimization problem and its kernelization we refer to the supplement and \cite{KloBreSonZie11}. 

While prior work on MKL imposes a $1$-norm constraint on the mixing coefficients to enforce sparse solutions lying on a standard simplex \cite{Bach04,SonRaeSchSch06,Zien07,RakBacCanGra07}, we employ a generalized $\ell_p$-norm constraint $\|\vbeta\|_p\leq1$ for $p\geq 1$ as used in \cite{KloBreSonZieLasMue09,KloBreSonZie11}. 
The implications of this modification in the context of image concept classification will be discussed throughout this paper.

%-------------------------------
\subsection{Kernel Target Alignment} \label{kta1}
%-------------------------------

The kernel alignment introduced by \cite{Cristianini02} measures the similarity of two matrices as a cosine angle of vectors under the Frobenius product
\begin{align}
  {\cal A}(K_1,K_2) := \frac{\langle K_1, K_2 \rangle_F}
      {\| K_1 \|_F \|K_2 \|_F},
  \label{eq:alignment1}
\end{align}
It was argued in \cite{twoStage} that centering is required in order to correctly reflect the test errors from SVMs via kernel alignment.  %Note that SVMs are invariant to centering due to their dual constraint $\sum_i \alpha_i y_i=0$.
Centering in the corresponding feature spaces \cite{1195996} can be achieved by taking the product $H K H$, with
\begin{equation}
  H := I - \frac{1}{n} {\bf 1} {\bf 1}^\top,
\end{equation}
$I$ is the identity matrix of size $n$ and $\bf 1$ is the column vector with all ones. The centered kernel which achieves a perfect separation of two classes is proportional to $\widetilde{\boldsymbol y}
\widetilde{\boldsymbol y}^\top$, where
\begin{equation}   \widetilde{\boldsymbol y} = (\widetilde{y}_i), \quad   \widetilde{y}_i := \left\{  \begin{array}{ll}   \frac{1}{n_+} & y_i = +1 \\ - \frac{1}{n_-} & y_i = -1 \end{array} \right.\label{eq:kta_idealsimilarity}\end{equation}
and $n_+$ and $n_-$ are the sizes of the positive and negative classes, respectively.

%%%%%%%%%%%%%%%%%%%%%%%%%%%%%%%%%%%%%%%%%%%%%%%%%%%%%%%%%%%%%%%%%%%%%%%%%%%%%%%%%%%%%%%%%%
\section{Empirical Evaluation}\label{experiment}
%%%%%%%%%%%%%%%%%%%%%%%%%%%%%%%%%%%%%%%%%%%%%%%%%%%%%%%%%%%%%%%%%%%%%%%%%%%%%%%%%%%%%%%%%%

In this section, we evaluate $\ell_p$-norm MKL in real-world image categorization tasks, experimenting on the VOC2009 and ImageCLEF2010 data sets. We also provide insights on \emph{when} and  \emph{why} $\ell_p$-norm MKL can help performance in image classification applications. The evaluation measure for both datasets is the average precision (AP) over all recall values based on the precision-recall (PR) curves.

%-------------------------------
\subsection{Data Sets}\label{VOCdataset}
%-------------------------------

We experiment on the following data sets:

\smallparagraph{1. PASCAL2 VOC  Challenge 2009}
We use the official data set of the  \emph{PASCAL2 Visual Object Classes Challenge 2009} (VOC2009) \cite{pascal-voc-2009}, which consists of 13979 images. The use the official split into 3473 training, 3581 validation, and 6925 test examples as provided by the challenge organizers.  The organizers also provided annotation of the 20 objects categories; note that an image can have multiple object annotations.  The task is to solve 20 binary classification problems, i.e. predicting whether at least one object from a class $k$ is visible in the test image. Although the test labels are undisclosed, the more recent VOC datasets permit to evaluate AP scores on the test set via the challenge website (the number of allowed submissions per week being limited).

\smallparagraph{2. ImageCLEF 2010 PhotoAnnotation}
The ImageCLEF2010 PhotoAnnotation  data set \cite{imageclef2010} consists of 8000 labeled training images taken from flickr and a test set with undisclosed labels. The images are annotated by 93 concept classes having highly variable concepts---they contain both well defined objects such as \emph{lake, river, plants, trees, flowers,} as well as many rather ambiguously defined concepts such as \emph{winter, boring, architecture, macro, artificial, motion blur,}---however, those concepts might not always be connected to objects present in an image or captured by a bounding box. This makes it highly challenging for any recognition system. Unfortunately, there is currently no official way to obtain test set performance scores from the challenge organizers. Therefore, for this data set, we report on training set cross-validation performances only. As for VOC2009 we decompose the problem into 93 binary classification problems. Again, many concept classes are challenging to rank or classify by an object detection approach due to their inherent non-object nature. As for the previous dataset each image can be labeled with multiple concepts.

%-------------------------------
\subsection{Image Features and Base Kernels}\label{basekernels}
%-------------------------------

In all of our experiments we deploy 32 kernels capturing various aspects of the images. The kernels are inspired by the VOC 2007 winner \cite{Marszalek07} and our own experiences from our submissions to the VOC2009 and ImageCLEF2009 challenges. We can summarize the employed kernels by the following three types of basic features:
\begin{itemize}
  \item Histogram over a bag of visual words over SIFT features (BoW-S), 15 kernels
  \item Histogram over a bag of visual words over color intensity histograms (BoW-C), 8 kernels
  \item Histogram of oriented gradients (HoG), 4 kernels
  \item Histogram of pixel color intensities (HoC), 5 kernels.
\end{itemize}

We used a higher fraction of bag-of-word-based features as we knew from our challenge submissions that they have a better performance than global histogram features. The intention was, however, to use a variety of different feature types that have been proven to be effective on the above datasets in the past---but at the same time obeying memory limitations of maximally 25GB per job as required by  computer facilities used in our experiments (we used a cluster of 23 nodes having in total 256 AMD64 CPUs and with memory limitations ranging in 32--96 GB RAM per node).

The above features are derived from histograms that contain \emph{no} spatial information. We therefore enrich the respective representations by using spatial tilings $1\times1,3\times1,2\times2,4\times4,8\times8$, which correspond to single levels of the pyramidal approach \cite{Lazebnik06} (this is for capturing the spatial context of an image). Furthermore, we apply a $\chi^2$ kernel on top of the enriched histogram features, which is an established kernel for capturing histogram features \cite{Zhang07}. The bandwidth of the $\chi^2$ kernel is thereby heuristically chosen as the mean $\chi^2$ distance over all pairs of training examples \cite{Lampert08}. 

The BoW features were constructed in a standard way \cite{Csurka04}: at first, the SIFT descriptors \cite{Lowe04} were calculated on a regular grid with 6 pixel pitches for each image, learning  a code book  of size $4000$ for the SIFT features and of size $900$ for the color histograms by $k$-means clustering (with a random initialization). Finally, all SIFT descriptors  were assigned to visual words (so-called \emph{prototypes}) and then summarized into histograms within entire images or sub-regions. We  computed the SIFT features over the following color combinations, which are inspired by the winners of the Pascal VOC 2008 challenge winners  from the university of Amsterdam \cite{vandeSandeTPAMI2010}: red-green-blue (RGB), normalized RGB, gray-opponentColor1-opponentColor2, and  gray-normalized OpponentColor1-OpponentColor2; in addition, we also use a simple gray channel.

We computed the 15-dimensional local color histograms over the color combinations red-green-blue, gray-opponentColor1-opponentColor2, gray, and hue (the latter being weighted by the pixel value of the value component in the HSV color representation). 

This means, for BoW-S, we considered five color channels with three spatial tilings each ($1\times1$, $3\times1$, and $2\times2$), resulting in 15 kernels; for BoW-C, we considered four color channels with two spatial tilings each ($1\times1$ and $3\times1$), resulting in 8 kernels.

The HoG features were computed by discretizing the orientation of the gradient vector at each pixel into 24 bins and then summarizing the discretized orientations into histograms within image regions \cite{Dalal05}. Canny detectors \cite{Canny86} are used to discard contributions from pixels, around which the image is almost uniform. We computed them over the color combinations red-green-blue,  gray-opponentColor1-opponentColor2, and gray, thereby using the two spatial tilings $4\times4$ and $8\times8$. For the experiments we used four kernels: a product kernel created from the two kernels with the red-green-blue color combination but using different spatial tilings, another product kernel created in the same way but using the  gray-opponentColor1-opponentColor2  color combination, and the two kernels using the gray channel alone (but differing in their spatial tiling).

The HoC features were constructed by discretizing pixel-wise color values and computing their 15 bin histograms within image regions. To this end, we used the color combinations red-green-blue, gray-opponentColor1-opponentColor2, and gray. For each color combination the spatial tilings $2\times2$, $3 \times 1$, and $4\times4$ were tried. In the experiments we deploy five kernels: a product kernel created from the three kernels with different spatial tilings with colors red-green-blue, a product kernel created from the three kernels with color combination gray-opponentColor1-opponentColor2, and the three kernels using the gray channel alone(differing in their spatial tiling).

Note that building a product kernel out of $\chi^2$ kernels boils down to concatenating feature blocks (but using a separate kernel width for each feature block). The intention here was to use single kernels at separate spatial tilings for the weaker features (for problems depending on a certain tiling resolution) and combined kernels with all spatial tilings merged into one kernel to keep the memory requirements low and let the algorithms select the best choice.  

In practice, the normalization of kernels is as important for MKL as the normalization of features is for training regularized linear or single-kernel models. This is owed to the bias 
introduced by the regularization: optimal feature / kernel weights are requested to be small, implying a bias to towards excessively up-scaled kernels.
In general, there are several ways of normalizing kernel functions. We apply the following normalization method, proposed in \cite{ZieOng07,ChaRak08} and entitled \emph{multiplicative normalization} in \cite{KloBreSonZie11}; on the feature-space level this normalization corresponds to rescaling training examples to unit variance,
\begin{align}\label{normalization}
    K \leftarrow \frac{1}{n}\text{tr} (K) - \frac{1}{n^2} {\bf1}^{\T} K {\bf1}. 
\end{align}

%-------------------------------
\subsection{Experimental Setup}\label{EXPSeT}
%-------------------------------

We treat the multi-label data set as binary classification problems, that is, for each object category we trained a one-vs.-rest classifier. Multiple labels per image render multi-class methods inapplicable as these require mutually exclusive labels for the images. The respective SVMs are trained using the Shogun toolbox \cite{SonRaeHenWidBehZieBonBinGehFra10new}. In order to shed light on the nature of the presented techniques from a statistical viewpoint, we first pooled all labeled data and then created 20 random cross-validation splits for VOC2009 and 12 splits for the larger dataset ImageCLEF2010. 

For each of the 12 or 20 splits, the training images were used for learning the classifiers, while the SVM/MKL regularization parameter $C$ and the norm parameter $p$ were chosen based on the maximal AP score on the validation images.  Thereby, the regularization constant $C$ is optimized by class-wise grid search over $C\in\{ 10^i\thinspace|\thinspace i=-1,-0.5,0,0.5,1\}$. Preliminary runs indicated that this way the optimal solutions are attained inside the grid. Note that for $p=\infty$ the $\ell_p$-norm MKL boils down to a simple SVM using a uniform kernel combination (subsequently called sum-kernel SVM). In our experiments, we used the average kernel SVM instead of the sum-kernel one. This is no limitation in this as both lead to identical result for an appropriate choice of the SVM regularization parameter.

For a rigorous evaluation, we would have to construct a separate codebook for each cross validation split. However, creating codebooks and assigning descriptors to visual words is a time-consuming process. Therefore, in our experiments we resort to the common practice of using a single codebook created from all training images contained in the official split. Although this could result in a slight overestimation of the AP scores, this affects all methods equally and does not favor any classification method more than another---our focus lies on a \textit{relative} comparison of the different classification methods; therefore there is no loss in exploiting this computational shortcut.

%-------------------------------
\subsection{Results}\label{results}
%-------------------------------

\begin{table*}[p]
  \small
  \centering
  \caption{\label{tab:VOC2009traincv}\small Average AP scores on the VOC2009 data set with AP scores computed by cross-validation on the training set. Bold faces show the best method and all other ones that are not statistical-significantly worse. }
  \vskip 0.05in
  \begin{tabular}{c c  c c c c c c } \hline  
   Norm & \textbf{Average}  &Aeroplane &Bicycle &Bird &Boat &Bottle &Bus \\ \hline  
   $\ell_1$  & 54.94 $\pm$ 12.3 & \textbf{84.84} $\pm$ 5.86 & 55.35 $\pm$ 10.5 & 59.38 $\pm$ 10.1 & 66.83 $\pm$ 12.4 & 25.91 $\pm$ 10.2 & \textbf{71.15} $\pm$ 23.2 \\  
   $\ell_{1.125}$  & 57.07 $\pm$ 12.7 & \textbf{84.82} $\pm$ 5.91 & \textbf{57.25} $\pm$ 10.6 & 62.4 $\pm$ 9.13 & \textbf{67.89} $\pm$ 12.8 & \textbf{27.88} $\pm$ 9.91 & \textbf{71.7} $\pm$ 22.8 \\  
 $\ell_{1.333}$  & 57.2 $\pm$ 12.8 & \textbf{84.51} $\pm$ 6.27 & \textbf{57.41} $\pm$ 10.8 & \textbf{62.75} $\pm$ 9.07 & \textbf{67.99} $\pm$ 13 & \textbf{27.44} $\pm$ 9.77 & \textbf{71.33} $\pm$ 23.1 \\  
 $\ell_{2}$  & 56.53 $\pm$ 12.8 & 84.12 $\pm$ 5.92 & 56.89 $\pm$ 10.9 & \textbf{62.53} $\pm$ 8.9 & 67.69 $\pm$ 13 & 26.68 $\pm$ 9.94 & 70.33 $\pm$ 22.3 \\  
 $\ell_{\infty}$  & 56.08 $\pm$ 12.7 & 83.67 $\pm$ 5.99 & 56.09 $\pm$ 10.9 & 61.91 $\pm$ 8.81 & 67.52 $\pm$ 12.9 & 26.5 $\pm$ 9.5 & 70.13 $\pm$ 22.2 \\  
 \hline  
 \hline  
 Norm & Car &Cat &Chair &Cow &Diningtable &Dog & Horse  \\ \hline  
 $\ell_1$  &  54.54 $\pm$ 7.33 & 59.5 $\pm$ 8.22 & 53.3 $\pm$ 11.7 & 23.13 $\pm$ 13.2 & \textbf{48.51} $\pm$ 19.9 & 41.72 $\pm$ 9.44 & 57.67 $\pm$ 12.2 \\  
 $\ell_{1.125}$  & \textbf{56.59} $\pm$ 8.93 & \textbf{61.59} $\pm$ 8.26 & \textbf{54.3} $\pm$ 12.1 & \textbf{29.59} $\pm$ 16.2 & \textbf{49.32} $\pm$ 19.5 & \textbf{45.57} $\pm$ 10.6 &  \textbf{59.4} $\pm$ 12.2 \\  
 $\ell_{1.333}$  &  \textbf{56.75} $\pm$ 9.28 & \textbf{61.74} $\pm$ 8.41 & \textbf{54.25} $\pm$ 12.3 & \textbf{29.89} $\pm$ 15.8 & 48.4 $\pm$ 19.3 & \textbf{45.85} $\pm$ 10.9 &  \textbf{59.4} $\pm$ 11.9 \\  
 $\ell_{2}$  & 55.92 $\pm$ 9.49 & 61.39 $\pm$ 8.37 & \textbf{53.85} $\pm$ 12.4 & 28.39 $\pm$ 16.2 & 47 $\pm$ 18.7 & 45.14 $\pm$ 10.8 &  58.61 $\pm$ 11.9\\  
 $\ell_{\infty}$  &  55.58 $\pm$ 9.47 & \textbf{61.25} $\pm$ 8.28 & 53.13 $\pm$ 12.4 & 27.56 $\pm$ 16.2 & 46.29 $\pm$ 18.8 & 44.63 $\pm$ 10.6 & 58.32 $\pm$ 11.7\\  
 \hline  
\hline  
 Norm & Motorbike &Person &Pottedplant &Sheep &Sofa  &Train &Tvmonitor \\ \hline  
 $\ell_1$    & 55 $\pm$ 13.2 & 81.32 $\pm$ 9.49 & 35.14 $\pm$ 13.4 & 38.13 $\pm$ 19.2 & \textbf{48.15} $\pm$ 11.8 & \textbf{75.33} $\pm$ 14.1 & 63.97 $\pm$ 10.2 \\  
 $\ell_{1.125}$   & \textbf{57.66} $\pm$ 13.1 & \textbf{82.18} $\pm$ 9.3 & 39.05 $\pm$ 14.9 & \textbf{43.65} $\pm$ 20.5 & \textbf{48.72} $\pm$ 13 & \textbf{75.79} $\pm$ 14.4 & \textbf{65.99} $\pm$ 9.83\\  
 $\ell_{1.333}$   & \textbf{57.57} $\pm$ 13 & \textbf{82.27} $\pm$ 9.29 & \textbf{39.7} $\pm$ 14.6 & \textbf{46.28} $\pm$ 23.9 & \textbf{48.76} $\pm$ 11.9 & \textbf{75.75} $\pm$ 14.3 & \textbf{66.07} $\pm$ 9.59 \\  
 $\ell_{2}$   & \textbf{56.9} $\pm$ 13.2 & 82.19 $\pm$ 9.3 & 38.97 $\pm$ 14.8 & \textbf{45.88} $\pm$ 24 & 47.29 $\pm$ 11.7  & \textbf{75.29} $\pm$ 14.5 & \textbf{65.55} $\pm$ 10.1 \\  
 $\ell_{\infty}$  & \textbf{56.45} $\pm$ 13.1 & 82 $\pm$ 9.37 & 38.46 $\pm$ 14.1 & \textbf{45.93} $\pm$ 24 & 46.08 $\pm$ 11.8  & 74.89 $\pm$ 14.5 & \textbf{65.19} $\pm$ 10.2 \\  
 \hline  
 \end{tabular}
\end{table*}

\begin{table*}[p]
  \centering
  \small
  \caption{\label{tab:VOC2009test}\small AP scores attained on the VOC2009 test data, obtained on request from the challenge organizers.  Best methods are marked boldface.}
  \vskip 0.05in
  \begin{tabular}{ccccccccc}
  \hline
 & \textbf{average} & aeroplane & bicycle & bird & boat & bottle & bus & car\\ \hline
$\ell_1$        & 54.58 	& \textbf{81.13}& 54.52 & 	56.14 & 	62.44 & \textbf{28.10} & \textbf{68.92} & 52.33 \\
$\ell_{1.125}$  & 56.43 	& 81.01 	& 56.36 & 	58.49 & 	62.84 & 25.75 & 68.22 & 55.71 \\
$\ell_{1.333}$    & \textbf{56.70} & 80.77 & \textbf{56.79} & \textbf{58.88} & 63.11 & 25.26 & 67.80 & \textbf{55.98} \\
$\ell_{2}$      & 56.34 	& 80.41 	& 56.34 & 	58.72 & \textbf{63.13} & 24.55 & 67.70 & 55.54 \\
$\ell_{\infty}$ & 55.85 	& 79.80 	& 55.68 & 	58.32 & 	62.76 & 24.23 & 67.79 & 55.38 \\
\hline
\hline
 &  & cat & chair & cow & diningtable & dog & horse & motorbike\\
\hline
$\ell_1$        & & 55.50 & 52.22 & 36.17 & 45.84 & 41.90 & 61.90 & 57.58 \\
$\ell_{1.125}$  & & 57.79 & 53.66 & 40.77 & \textbf{48.40} & 46.36 & \textbf{63.10} & 60.89 \\
$\ell_{1.333}$    & & \textbf{58.00} & \textbf{53.87} & \textbf{43.14} & 48.17 & 46.54 & 63.08 & \textbf{61.28} \\
$\ell_{2}$      & & 57.98 & 53.47 & 40.95 & 48.07 & \textbf{46.59} & 63.02 & 60.91 \\
$\ell_{\infty}$ & & 57.30 & 53.07 & 39.74 & 47.27 &45.87 & 62.49 & 60.55 \\ 
\hline
\hline
 &  & person & pottedplant & sheep & sofa & train & tvmonitor &\\
\hline
$\ell_1$        & & 81.73 & 31.57 & 36.68 & 45.72 & \textbf{80.52} & 61.41       &\\
$\ell_{1.125}$  & & 82.65 & \textbf{34.61} & 41.91 & \textbf{46.59} & 80.13 & 63.51       &\\
$\ell_{1.333}$    & & \textbf{82.72} & 34.60 & 44.14 & 46.42 & 79.93 & \textbf{63.60}       &\\
$\ell_{2}$      & & 82.52 & 33.40 & \textbf{44.81} & 45.98 & 79.53 & 63.26       &\\
$\ell_{\infty}$ & & 82.20 & 32.76 & 44.15 & 45.69 & 79.03 & 63.00       &\\
\hline
\end{tabular}
\end{table*}

\begin{table*}[htp]
  \centering
  \small 
   \caption{\label{tab:voccwscores}\small Average AP scores on the VOC2009 data set with norm parameter $p$ class-wise optimized over AP scores on the training set. We report on test set scores obtained on request from the challenge organizers.}
  \vskip 0.05in
  \begin{tabular}{c c c c c c}
  \hline  $\infty$ & $\{1,\infty\}$ & $\{1.125,1.333,2\}$ & $\{1.125,1.333,2,\infty\} $ & $\{1,1.125,1.333,2\}$ & all norms from the left \\ \hline
    55.85  & 55.94  & 56.75 & 56.76 & 56.75 & 56.76 \\  \hline
  \end{tabular}
\end{table*}

\begin{table*}[htp]
  \small
  \centering
   \caption{\label{tab:clefsmallscores}\small Average AP scores obtained on the ImageCLEF2010 data set with $p$ fixed over the classes and AP scores computed by cross-validation on the training set.}
  \vskip 0.05in
  \begin{tabular}{c c c c c c}
  \hline
  $\ell_p$-Norm & 1 & 1.125 & 1.333 & 2 & $\infty$ \\ \hline
	  & 37.32 $\pm$ 5.87 & 39.51 $\pm$ 6.67 & 39.48 $\pm$ 6.66 & 39.13 $\pm$ 6.62 & 39.11 $\pm$ 6.68 \\  \hline
  \end{tabular}
\end{table*}

\begin{table*}[htp]
  \small
  \centering
  \caption{\label{tab:clefcwscores}\small Average AP scores obtained on the ImageCLEF2010 data set with norm parameter $p$ class-wise optimized and AP scores computed by cross-validation on the training set.}
  \vskip 0.05in
  \begin{tabular}{c c c c c c}
    \hline
    $\infty$ & $\{1,\infty\}$ & $\{1.125,1.333,2\}$ & $\{1.125,1.333,2,\infty\} $ & $\{1,1.125,1.333,2\}$ & all norms from the left \\ \hline
  	39.11 $\pm$ 6.68 & 39.33 $\pm$ 6.71 & 39.70 $\pm$ 6.80 & 39.74 $\pm$ 6.85 & 39.82 $\pm$ 6.82 & 39.85 $\pm$ 6.88 \\  \hline
  \end{tabular}
\end{table*}

In this section we report on the empirical results achieved by $\ell_p$-norm MKL in our visual object recognition experiments.

\smallparagraph{VOC 2009}
Table~\ref{tab:VOC2009test} shows the AP scores attained on the official test split of the VOC2009 data set (scores obtained by evaluation via the challenge website). The class-wise optimal regularization constant has been selected by cross-validation-based model selection on the training data set. We can observe that non-sparse MKL outperforms the baselines $\ell_1$-MKL and the sum-kernel SVM in this sound evaluation setup. We also report on the cross-validation performance achieved on the training data set (Table~\ref{tab:VOC2009traincv}). Comparing the two results, one can observe a small overestimation for the cross-validation approach (for the reasons argued in Section~\ref{EXPSeT})---however, the amount by which this happens is equal for all methods; in particular, the ranking of the compared methods (SVM versus $\ell_p$-norm MKL for various values of $p$) is preserved for the average over all classes and most of the classes (exceptions are the bottle and bird class); this shows the reliability of the cross-validation-based evaluation method in practice. Note that the observed variance in the AP measure across concepts can be explained in part by the variations in the label distributions across concepts and cross-validation splits. Unlike for the AUC measure, the average score of the AP measure under randomly ranked images depends on the ratio of positive and negative labeled samples.

A reason why the bottle class shows such a strong deviation towards sparse methods could be the varying but often small fraction of image area covered by bottles leading to overfitting when using spatial tilings. 

We can also remark that $\ell_{1.333}$-norm achieves the best result of all compared methods on the VOC dataset, slightly followed by $\ell_{1.125}$-norm MKL. To evaluate the statistical significance of our findings, we perform a Wilcoxon signed-rank test for the cross-validation-based results  (see Table~\ref{tab:VOC2009traincv}; significant results are marked in boldface). We find that in 15 out of the 20 classes the optimal result is achieved by truly non-sparse $\ell_p$-norm MKL (which means $p\in]1,\infty[$), thus outperforming the baseline significantly.

\smallparagraph{ImageCLEF}
Table~\ref{tab:clefsmallscores} shows the AP scores averaged over all classes achieved on the ImageCLEF2010 data set. We observe that the best result is achieved by the non-sparse $\ell_p$-norm MKL algorithms with norm parameters $p=1.125$ and $p=1.333$. The detailed results for all 93 classes are shown in the supplemental material (see B.1~and~B.2.%Tables~\ref{tab:clefscorespart1}~and~\ref{tab:clefscorespart2}). 
We can see from the detailed results that in 37 out of the 93 classes the optimal result attained by non-sparse $\ell_p$-norm MKL was significantly better than the sum kernel according to a Wilcoxon signed-rank test.

We also show the results for optimizing the norm parameter $p$ \emph{class-wise} (see Table~\ref{tab:clefcwscores}). We can see from the table that optimizing the $\ell_p$-norm class-wise is beneficial: selecting the best $p\in]1,\infty[$ class-wise, the result is increased to an AP of 39.70. Also including $\ell_1$-norm MKL in the candidate set, the performance can even be leveraged to 39.82---this is 0.7 AP better than the result for the vanilla sum-kernel SVM. Also including the latter to the set of model, the AP score only merely increases by 0.03 AP points. 
We conclude that optimizing the norm parameter $p$ class-wise can improve performance; however, one can rely on $\ell_p$-norm MKL alone without the need to additionally include the sum-kernel-SVM to the set of models. Tables \ref{tab:VOC2009traincv} and \ref{tab:VOC2009test} show that the gain in performance for MKL varies considerably on the actual concept class. Notice that these observations are confirmed by the results presented in Tables B.1 and B.2, see supplemental material for details.

%-------------------------------
\subsection{Analysis and Interpretation}
%-------------------------------

We now analyze the kernel set in an explorative manner; to this end, our methodological tools are the following
\begin{enumerate}
\item Pairwise kernel alignment scores (KA)
\item Centered kernel-target alignment scores (KTA).
\end{enumerate}

\subsubsection{Analysis of the Chosen Kernel Set}

\begin{figure}[htp]
  \begin{center}
  \includegraphics[height=0.22\textwidth]{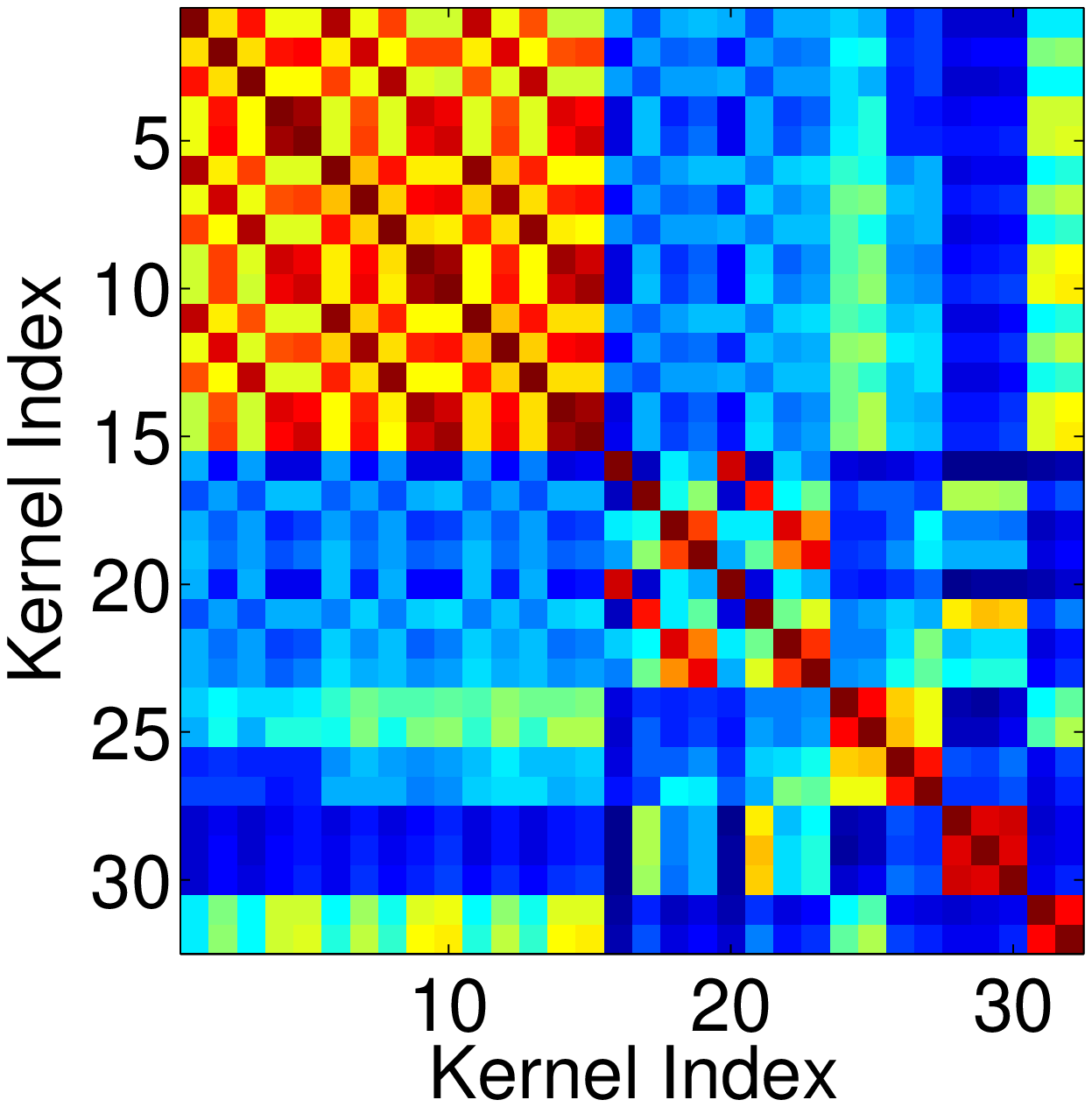} 
  \includegraphics[height=0.22\textwidth]{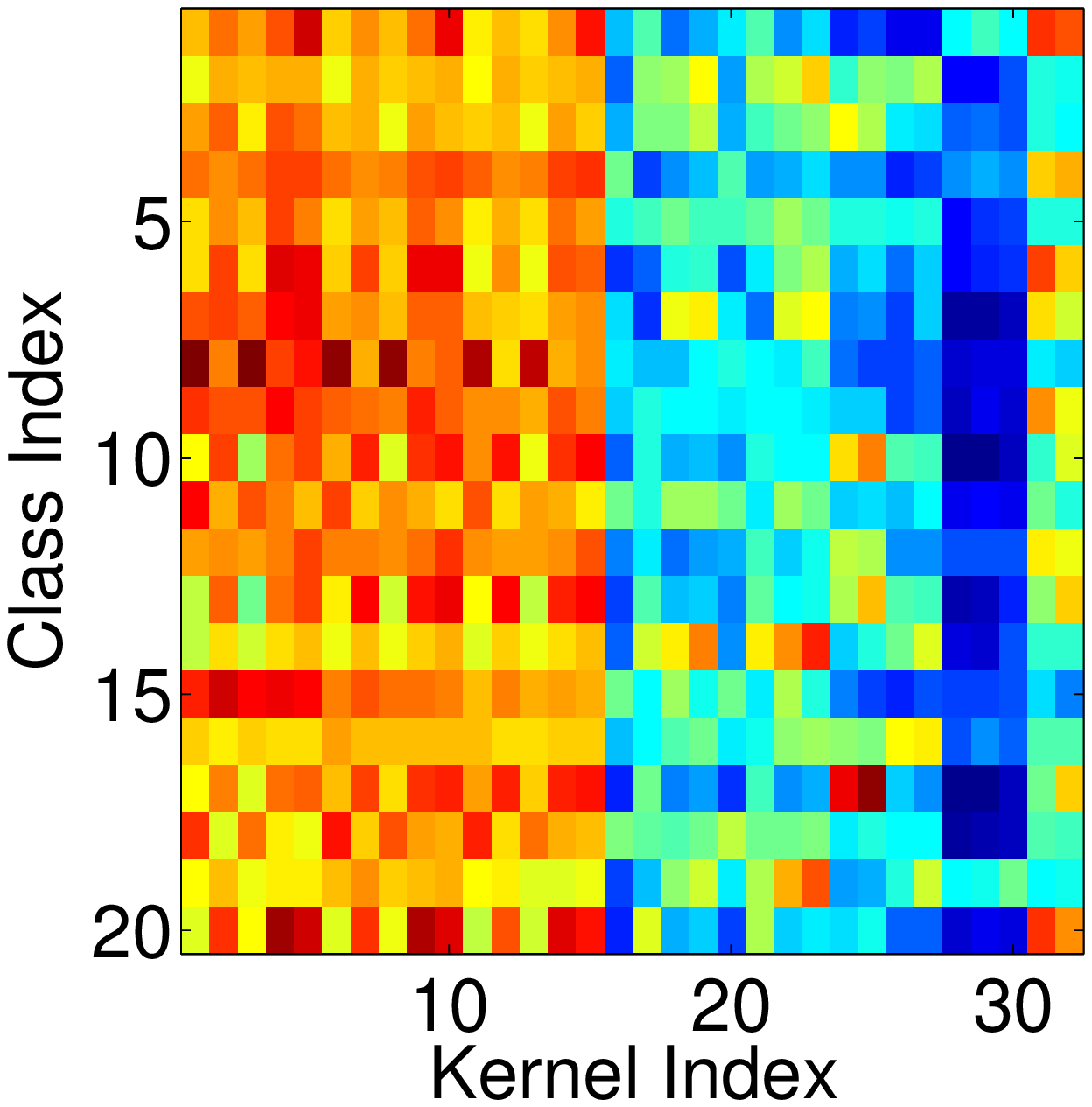}\\
  \includegraphics[height=0.22\textwidth]{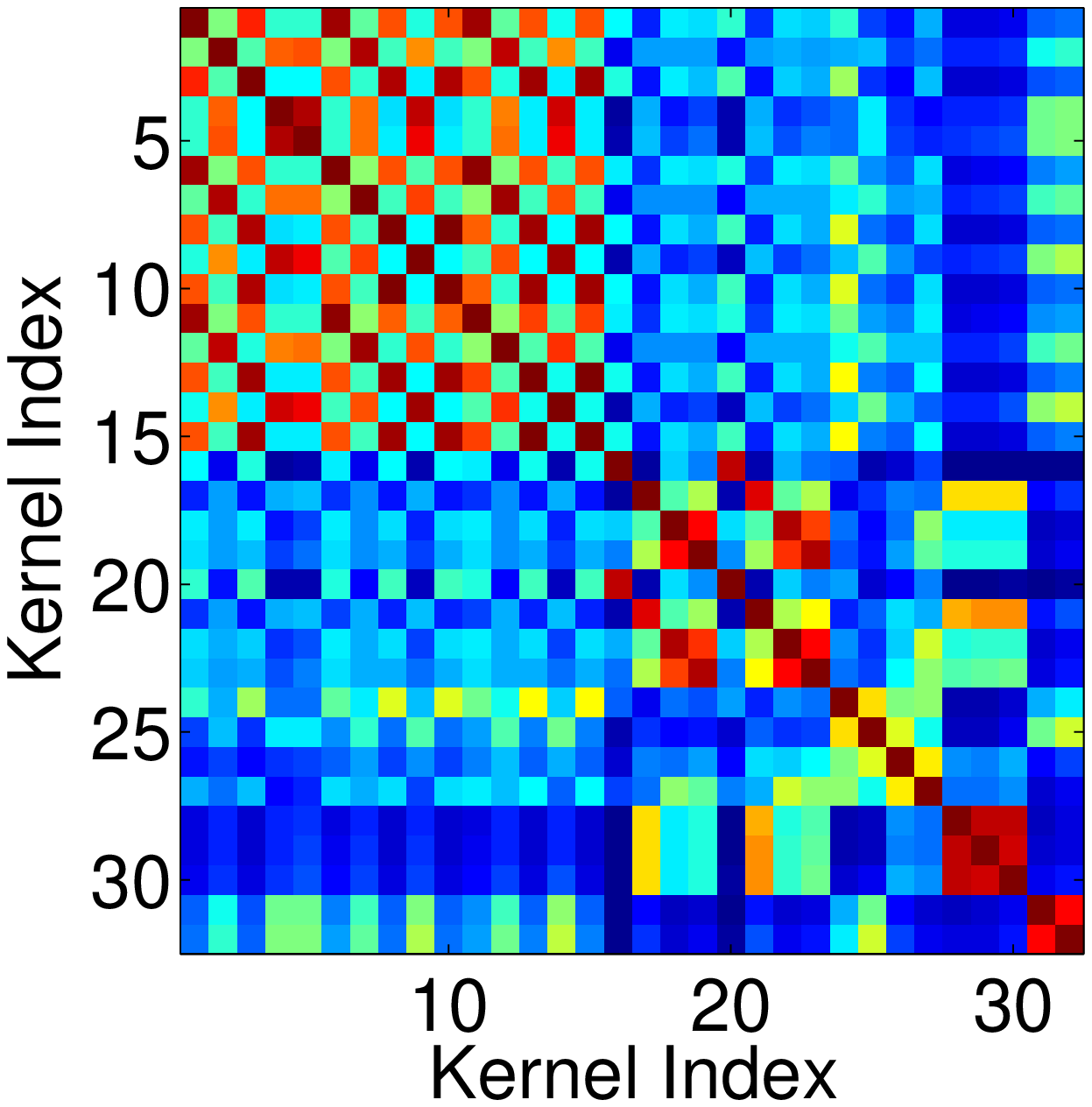}
  \includegraphics[height=0.22\textwidth]{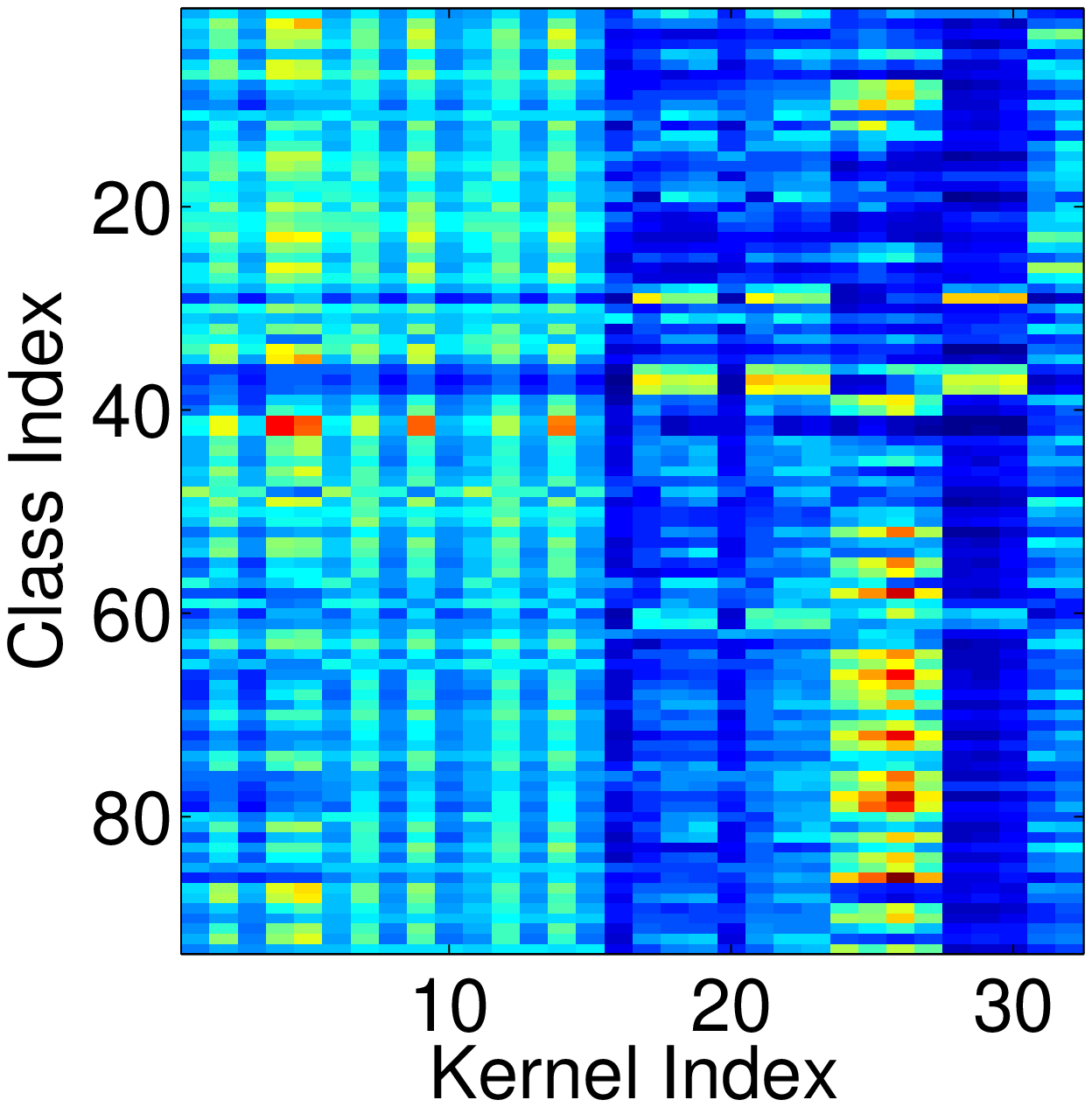}
  \end{center}
  \caption{\label{fig:kka}\small Similarity of the kernels for the VOC2009  (\textsc{Top}) and ImageCLEF2010 (\textsc{Bottom}) data sets in terms of pairwise kernel alignments (\textsc{Left}) and kernel target alignments (\textsc{Right}), respectively. In both data sets, five groups can be identified: 'BoW-S' (Kernels 1--15), 'BoW-C' (Kernels 16--23), 'products of HoG and HoC kernels' (Kernels 24--27, 'HoC single' (Kernels 28--30), and 'HoG single' (Kernels 31--32).}
\end{figure}

To start with, we computed the pairwise kernel alignment scores of the 32 base kernels: they are shown in Fig.~\ref{fig:kka}. 
We recall that the kernels can be classified into the following groups: Kernels 1--15 and 16--23 employ BoW-S and BoW-C features, respectively; Kernels 24 to 27 are product kernels associated with the HoG and HoC features; Kernels 28--30 deploy HoC, and, finally, Kernels 31--32 are based on HoG features over the gray channel. We see from the block-diagonal structure that features that are of the same type (but are generated for different parameter values, color channels, or spatial tilings) are strongly correlated. Furthermore the BoW-S kernels (Kernels 1--15) are weakly correlated with the BoW-C kernels (Kernels 16--23). Both, the BoW-S and HoG kernels (Kernels 24--25,31--32) use gradients and therefore are moderately correlated; the same holds for the BoW-C and HoC kernel groups (Kernels 26--30). This corresponds to our original intention to have a broad range of feature types which are, however, useful for the task at hand. The principle usefulness of our feature set can be seen a posteriori from the fact that $\ell_1$-MKL achieves the worst performance of all methods included in the comparison while the sum-kernel SVM performs moderately well. Clearly, a higher fraction of noise kernels would further harm the sum-kernel SVM and favor the sparse MKL instead (we investigate the impact of noise kernels on the performance of $\ell_p$-norm MKL in an experiment on controlled, artificial data; this is presented in the supplemental material.

Based on the observation that the BoW-S kernel subset shows high KTA scores, we also evaluated the performance restricted to the 15 BoW-S kernels only. Unsurprisingly, this setup favors the sum-kernel SVM, which achieves higher results on VOC2009 for most classes; compared to $\ell_p$-norm MKL using all 32 classes, the sum-kernel SVM restricted to 15 classes achieves slightly better AP scores for 11 classes, but also slightly worse for 9 classes. Furthermore, the sum kernel SVM, $\ell_2$-MKL, and $\ell_{1.333}$-MKL were on par with differences fairly below 0.01 AP. This is again not surprising as the kernels from the BoW-S kernel set are strongly correlated with each other for the VOC data which can be seen in the top left image in Fig.~\ref{fig:kka}. For the ImageCLEF data we observed a quite different picture: the sum-kernel SVM restricted to the 15 BoW-S kernels performed significantly worse, when, again, being compared to non-sparse $\ell_p$-norm MKL using all 32 kernels. To achieve top state-of-the-art performance, one could optimize the scores for both datasets by considering the class-wise maxima over learning methods \emph{and} kernel sets. However, since the intention here is not to win a challenge but a relative comparison of models, giving insights in the nature of the methods---we therefore discard the time-consuming optimization over the kernel subsets.

From the above analysis, the question arises why restricting the kernel set to the 15 BoW-S kernels affects the performance of the compared methods differently, for the VOC2009 and ImageCLEF2010 data sets. This can be explained by comparing the KA/KTA scores of the kernels attained on VOC and on ImageCLEF (see Fig.~\ref{fig:kka}~(\textsc{Right})): for the ImageCLEF data set the KTA scores are substantially more spread along all kernels; there is neither a dominance of the BoW-S subset in the KTA scores nor a particularly strong correlation within the BoW-S subset in the KA scores. We attribute this to the less object-based and more ambiguous nature of many of the concepts contained in the ImageCLEF data set. Furthermore, the KA scores  for the ImageCLEF data (see Fig.~\ref{fig:kka}~(\textsc{Left})) show that this dataset exhibits a higher variance among kernels---this is because the correlations between all kinds of kernels are weaker for the ImageCLEF data.

Therefore, because of this non-uniformity in the spread of the information content among the kernels, we can conclude that indeed our experimental setting falls into the situation where non-sparse MKL can outperform the baseline procedures (again, see suuplemental material. For example, the BoW features are more informative than HoG and HoC, and thus the uniform-sum-kernel-SVM is suboptimal. On the other hand, because of the fact that typical image features are only moderately informative, HoG and HoC still convey a certain amount of complementary information---this is what allows the performance gains reported in Tables~\ref{tab:VOC2009traincv}~and~\ref{tab:clefsmallscores}.

Note that we class-wise normalized the KTA scores to sum to one. This is because we are rather interested in a comparison of the relative contributions of the particular kernels than in their absolute information content, which anyway can be more precisely derived from the AP scores already reported in Tables~\ref{tab:VOC2009traincv}~and~\ref{tab:clefsmallscores}. Furthermore, note that we consider \emph{centered} KA and KTA scores, since it was argued in \cite{twoStage} that only those correctly reflect the test errors attained by established learners such as SVMs.

\subsubsection{The Role of the Choice of $\ell_p$-norm}
\label{ssec:rolepnorm}

\begin{figure*}[htp]
  \centering
  \includegraphics[width=0.4\textwidth]{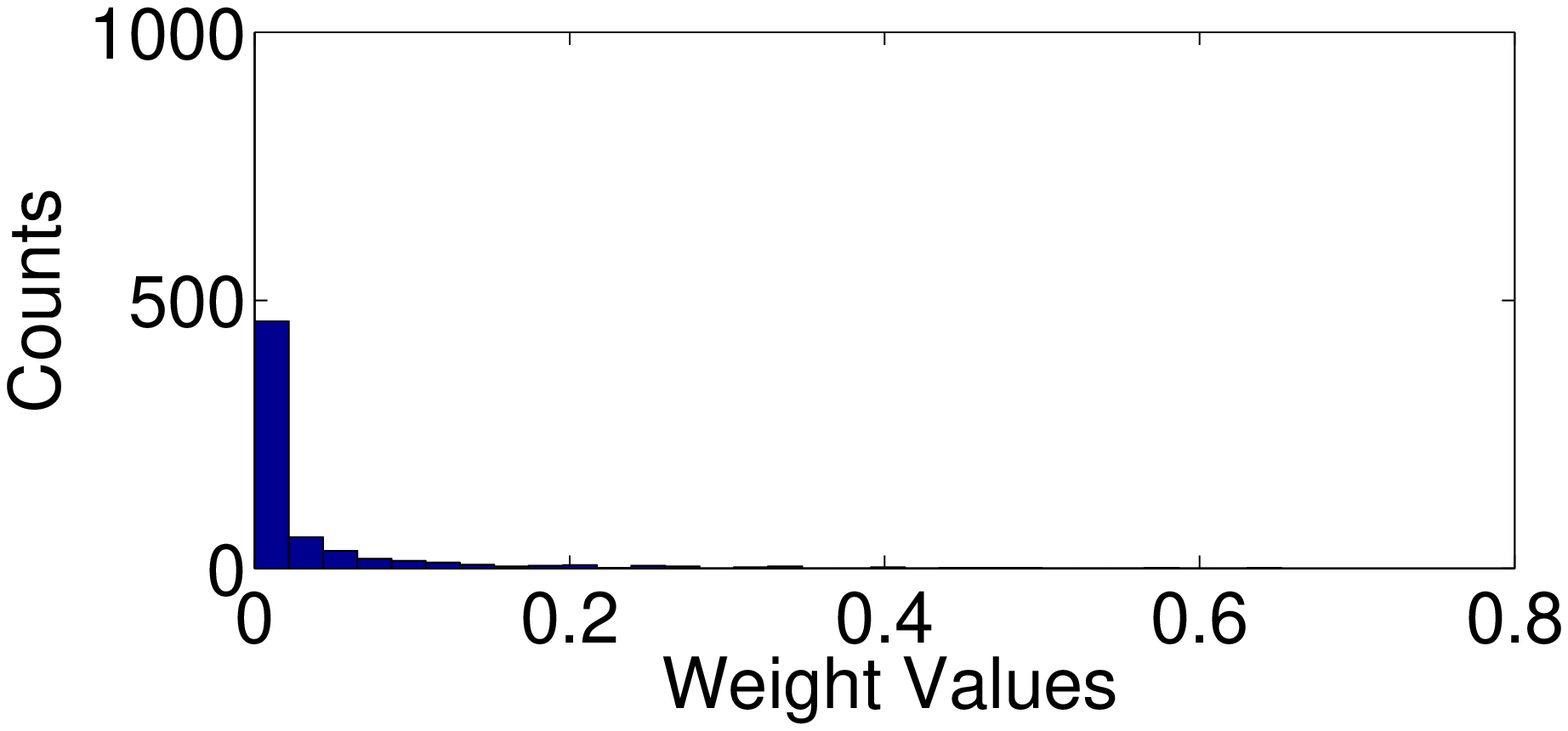} 
  \includegraphics[width=0.4\textwidth]{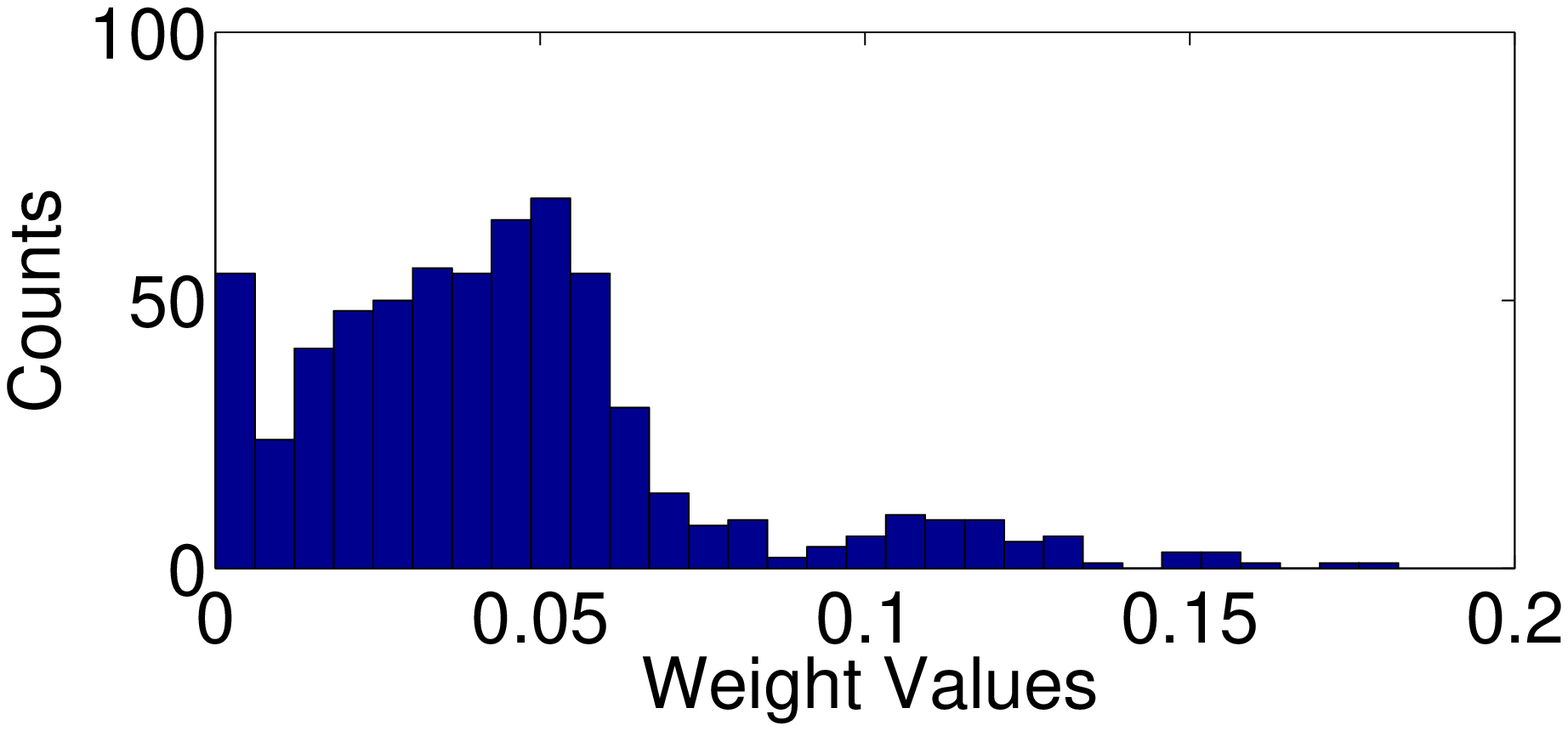} \\
  \includegraphics[width=0.4\textwidth]{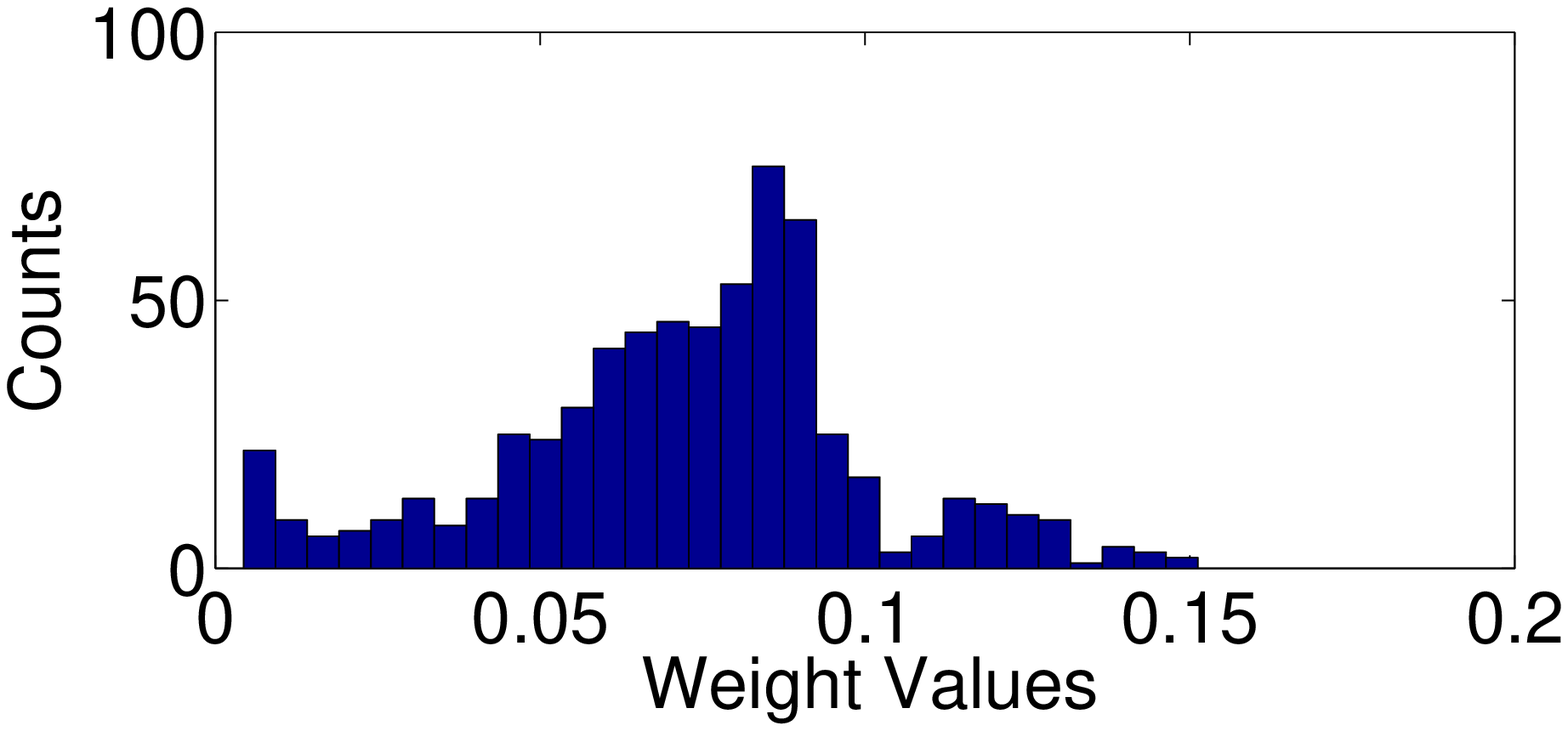} 
  \includegraphics[width=0.4\textwidth]{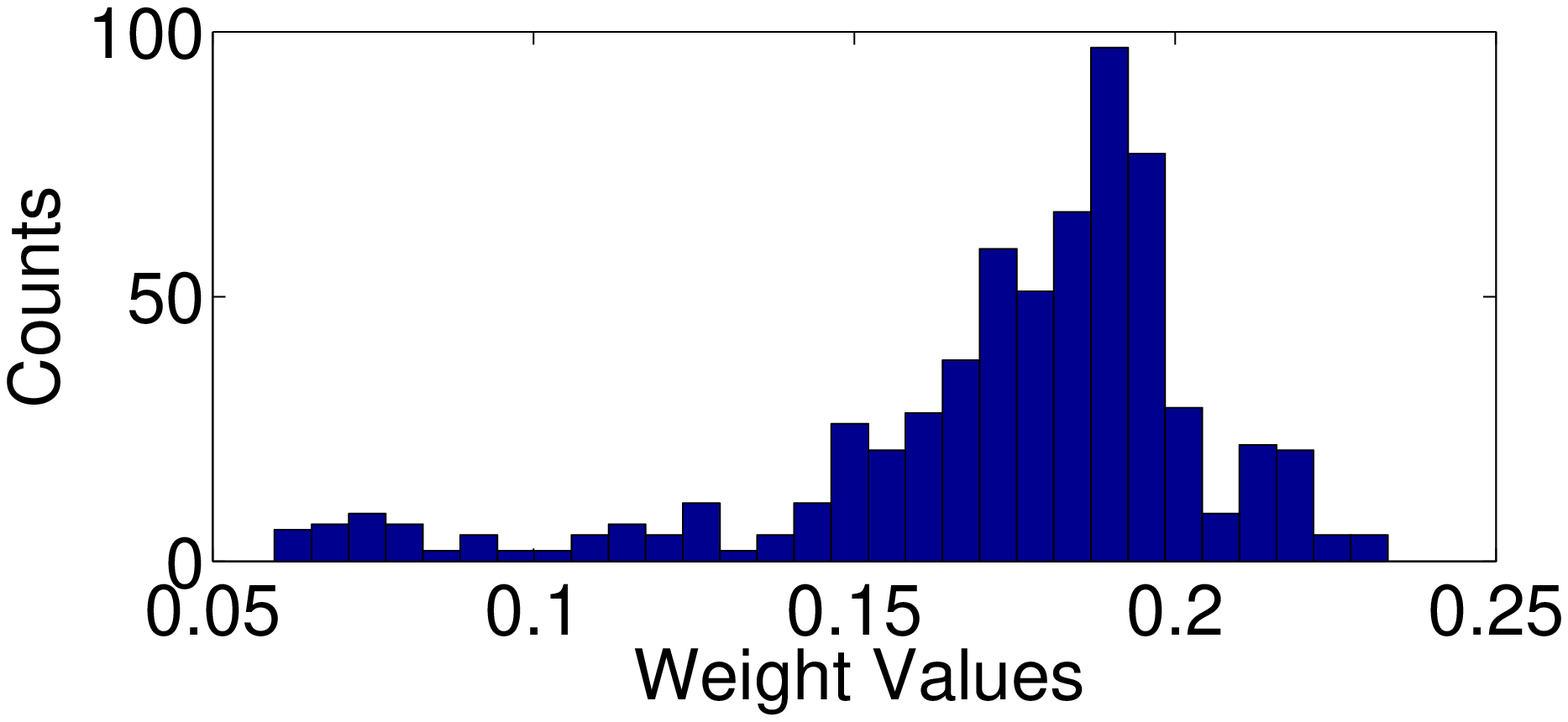} 
  \caption{\label{fig:kwhist_K32}\small Histogram of kernel weights as output by $\ell_p$-norm MKL for the various classes on the VOC2009 data set (32 kernels $\times$ 20 classes, resulting in 640 values):  $\ell_1$-norm (\textsc{top left})), $\ell_{1.125}$-norm (\textsc{top right}), $\ell_{1.333}$-norm (\textsc{bottom left}), and $\ell_{2}$-norm (\textsc{bottom right}).}
  \vspace{1mm}
\end{figure*}

Next, we turn to the interpretation of the norm parameter $p$ in our algorithm. We observe a big gap in performance between $\ell_{1.125}$-norm MKL and the sparse $\ell_{1}$-norm MKL. The reason is that for $p>1$ MKL is reluctant to set kernel weights to zero, as can be seen from Figure~\ref{fig:kwhist_K32}. In contrast, $\ell_{1}$-norm MKL eliminates 62.5\% of the kernels from the working set. The difference between the $\ell_p$-norms for $p>1$ lies solely in the ratio by which the less informative kernels are down-weighted---they are never assigned with true zeros.

However, as proved in \cite{KloBreSonZie11}, in the computational optimum, the kernel weights are accessed by the MKL algorithm via the information content of the particular kernels given by a  $\ell_p$-norm-dependent formula (see Eq.~\eqref{eq:mklinformation}; this will be discussed in detail in Section~\ref{ssec:argumentproavgkernel}). We mention at this point that the kernel weights all converge to the same, uniform value for $p\rightarrow\infty$. We can confirm these theoretical findings empirically: the histograms of the kernel weights shown in Fig.~\ref{fig:kwhist_K32} clearly indicate an increasing uniformity in the distribution of kernel weights when letting $p\rightarrow\infty$. Higher values of $p$ thus cause the weight distribution to shift away from zero and become slanted to the right while smaller ones tend to increase its skewness to the left.

Selection of the $\ell_p$-norm permits to tune the strength of the regularization of the learning of kernel weights. In this sense the sum-kernel SVM clearly is an extreme, namely fixing the kernel weights, obtained when letting $p\rightarrow\infty$. The sparse MKL marks another extreme case: $\ell_p$-norms with $p$ below $1$ loose the convexity property so that $p=1$ is the maximally sparse choice preserving convexity at the same time. Sparsity can be interpreted here that only a few kernels are selected which are considered most informative according to the optimization objective. Thus, the $\ell_p$-norm acts as a prior parameter for how much we trust in the informativeness of a kernel. In conclusion, this interpretation justifies the usage of $\ell_p$-norm outside the existing choices $\ell_1$ and $\ell_2$. The fact that the sum-kernel SVM is a reasonable choice in the context of image annotation will be discussed further in Section~\ref{ssec:argumentproavgkernel}.

Our empirical findings on ImageCLEF and VOC seem to contradict previous ones about the usefulness of MKL reported in the literature, where $\ell_1$ is frequently to be outperformed by a simple sum-kernel SVM (for example, see \cite{gehler_kernelfigureitout_cvpr2009})---however, in these studies the sum-kernel SVM is compared to \emph{$\ell_1$}-norm or \emph{$\ell_2$}-norm MKL only. In fact, our results \emph{confirm} these findings: $\ell_1$-norm MKL is outperformed by the sum-kernel SVM in all of our experiments. Nevertheless, in this paper, we show that by using the more general $\ell_p$-norm regularization, the prediction accuracy of MKL can be considerably leveraged, even clearly outperforming the sum-kernel SVM, which has been shown to be a tough competitor in the past \cite{GehlerN09}. But of course also the simpler sum-kernel SVM also has its advantage, although on the computational side only: in our experiments it was about a factor of ten faster than its MKL competitors. Further information about runtimes of MKL algorithms compared to sum kernel SVMs can be taken from \cite{KloftJMLR}.

\subsubsection{Remarks for Particular Concepts}

\begin{figure}[htp]
 \begin{center}
  \includegraphics[width=0.22\textwidth]{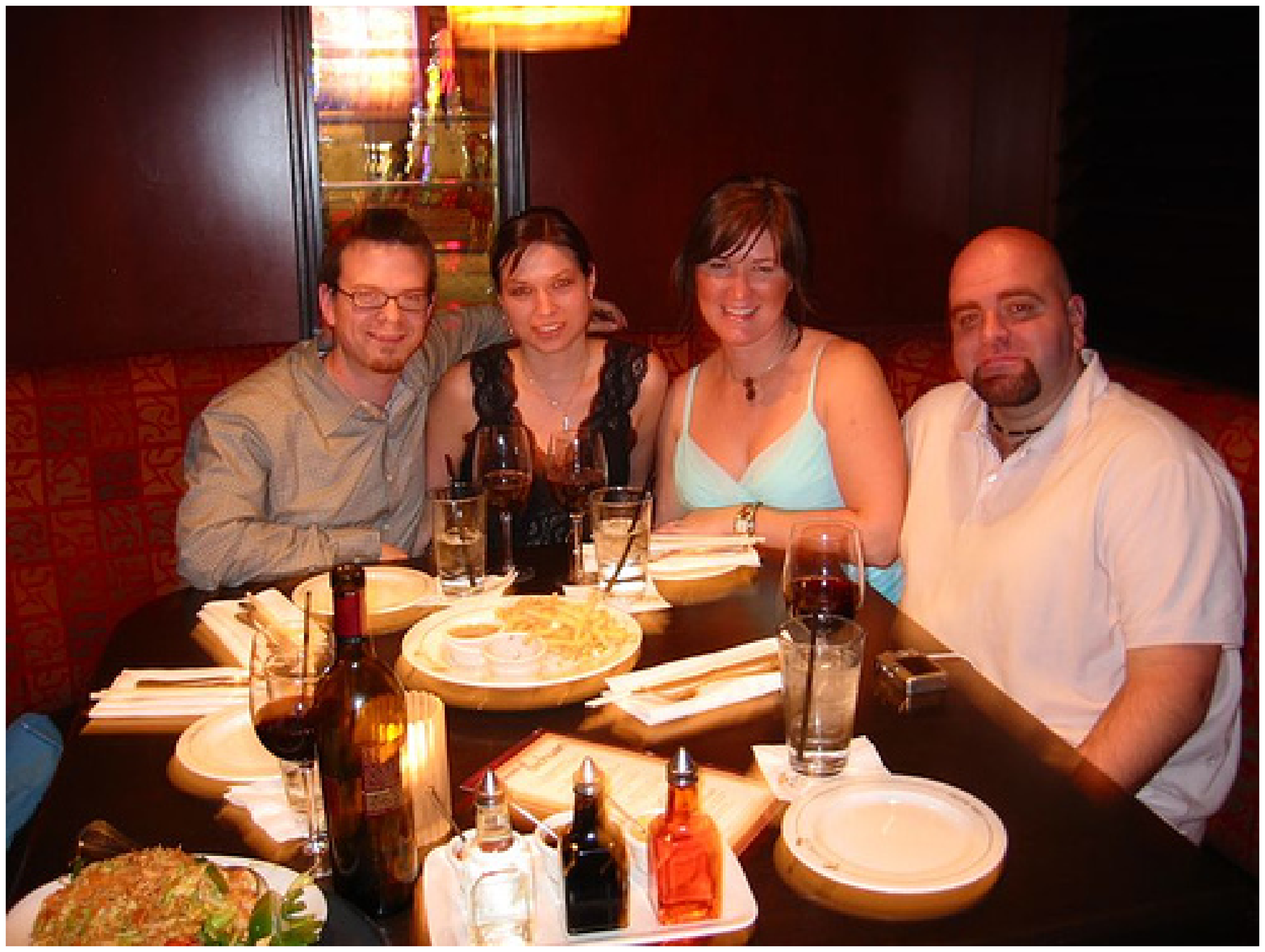} \hspace{0.02\textwidth}
  \includegraphics[width=0.22\textwidth]{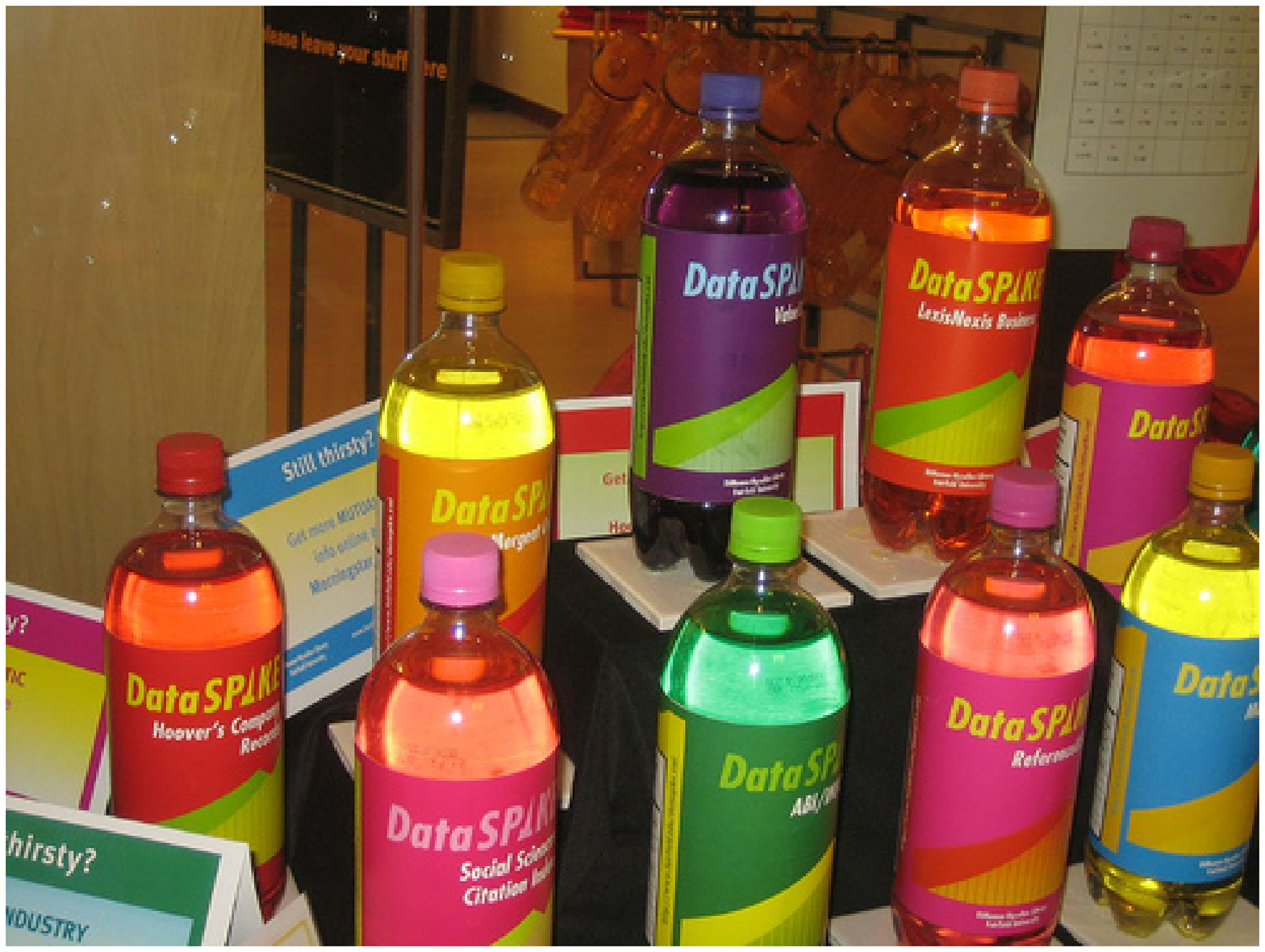} \\
  \vspace{5mm}
  \includegraphics[width=0.24\textwidth]{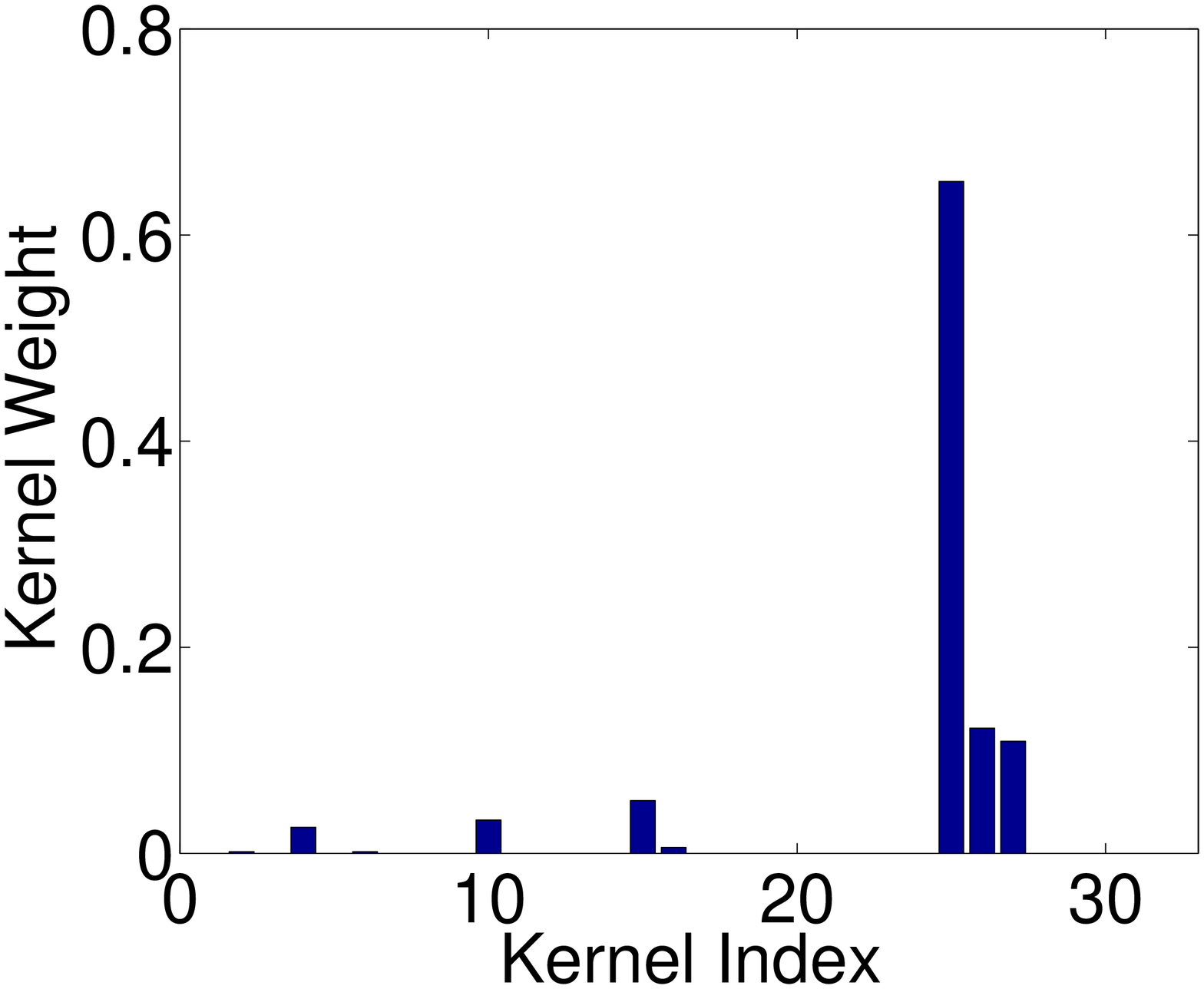} %\hspace{0.05\textwidth}
   \includegraphics[width=0.24\textwidth]{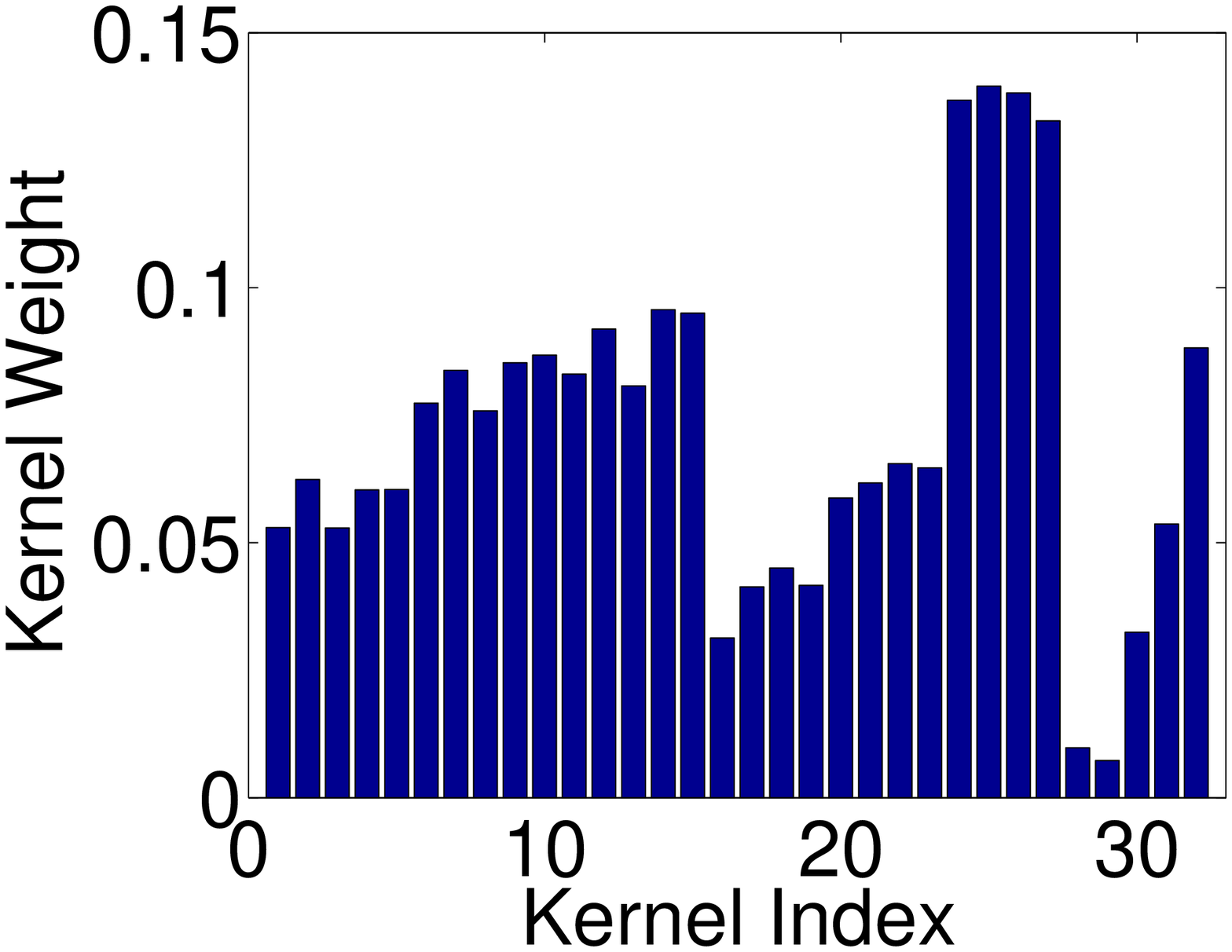}
  \caption{\small\label{fig:example_bottle} Images of typical highly ranked bottle images and kernel weights from $\ell_1$-MKL (left) and $\ell_{1.333}$-MKL (right).}
  \end{center}
\end{figure}

Finally, we show images from classes where MKL helps performance and discuss relationships to kernel weights. We have seen above that the sparsity-inducing $\ell_1$-norm MKL clearly outperforms all other methods on the \emph{bottle} class (see Table~\ref{tab:VOC2009test}). Fig.~\ref{fig:example_bottle} shows two typical highly ranked images and the corresponding kernel weights as output by $\ell_1$-norm (\textsc{Left}) and $\ell_{1.333}$-norm MKL (\textsc{Right}), respectively, on the bottle class. We observe that $\ell_1$-norm MKL tends to rank highly party and people group scenes. We conjecture that this has two reasons: first, many people group and party scenes come along with co-occurring bottles. Second, people group scenes have similar gradient distributions to images of large upright standing bottles sharing many dominant vertical lines and a thinner head section---see the left- and right-hand images in Fig.~\ref{fig:example_bottle}. Sparse $\ell_1$-norm MKL strongly focuses on the dominant HoG product kernel, which is able to capture the aforementioned special gradient distributions, giving small weights to two HoC product kernels and almost completely discarding all other kernels. 

\begin{figure}[htp]
 \begin{center}
  \includegraphics[width=0.22\textwidth]{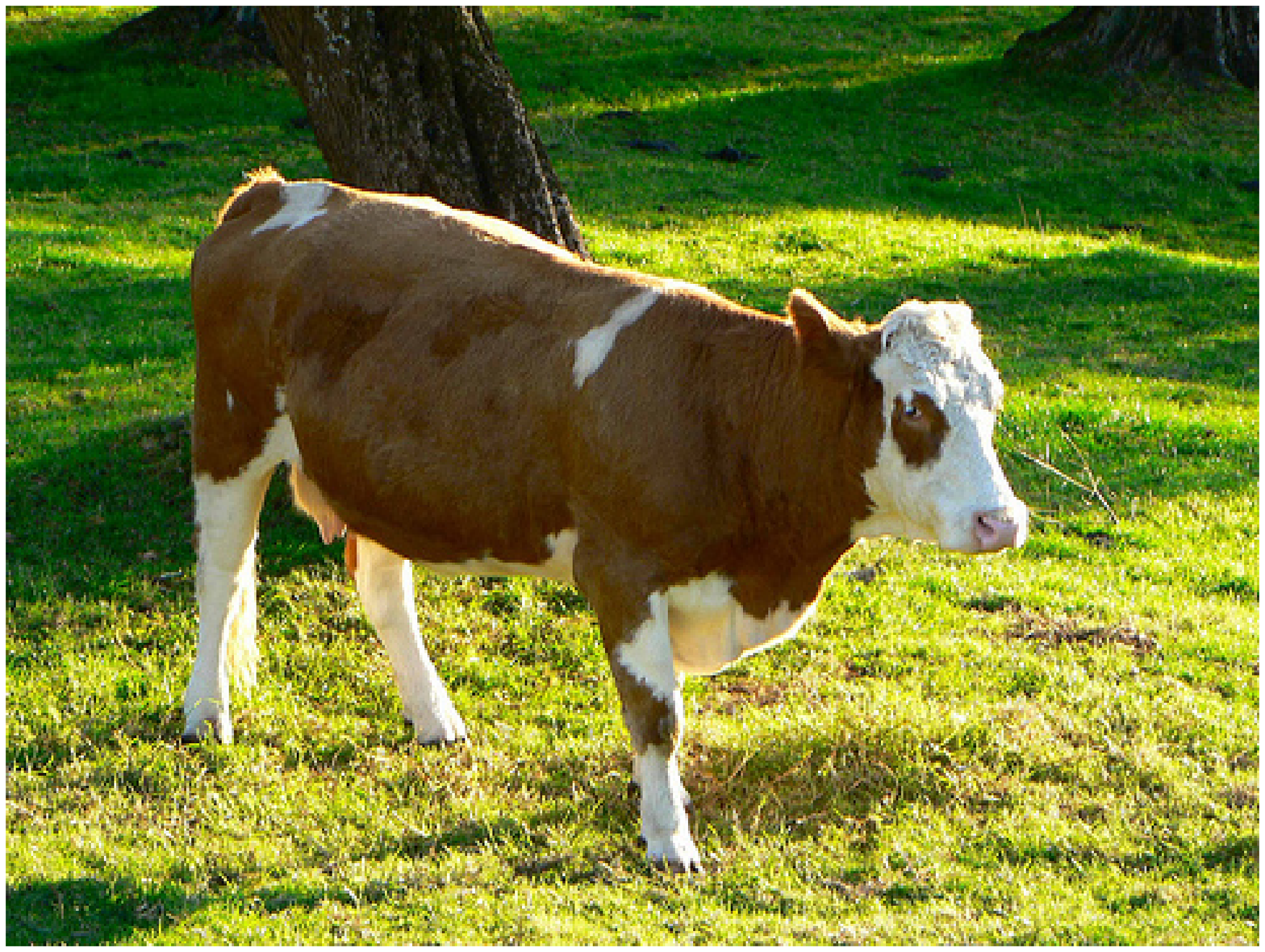} \\ 
  \vspace{5mm}
  \includegraphics[width=0.24\textwidth]{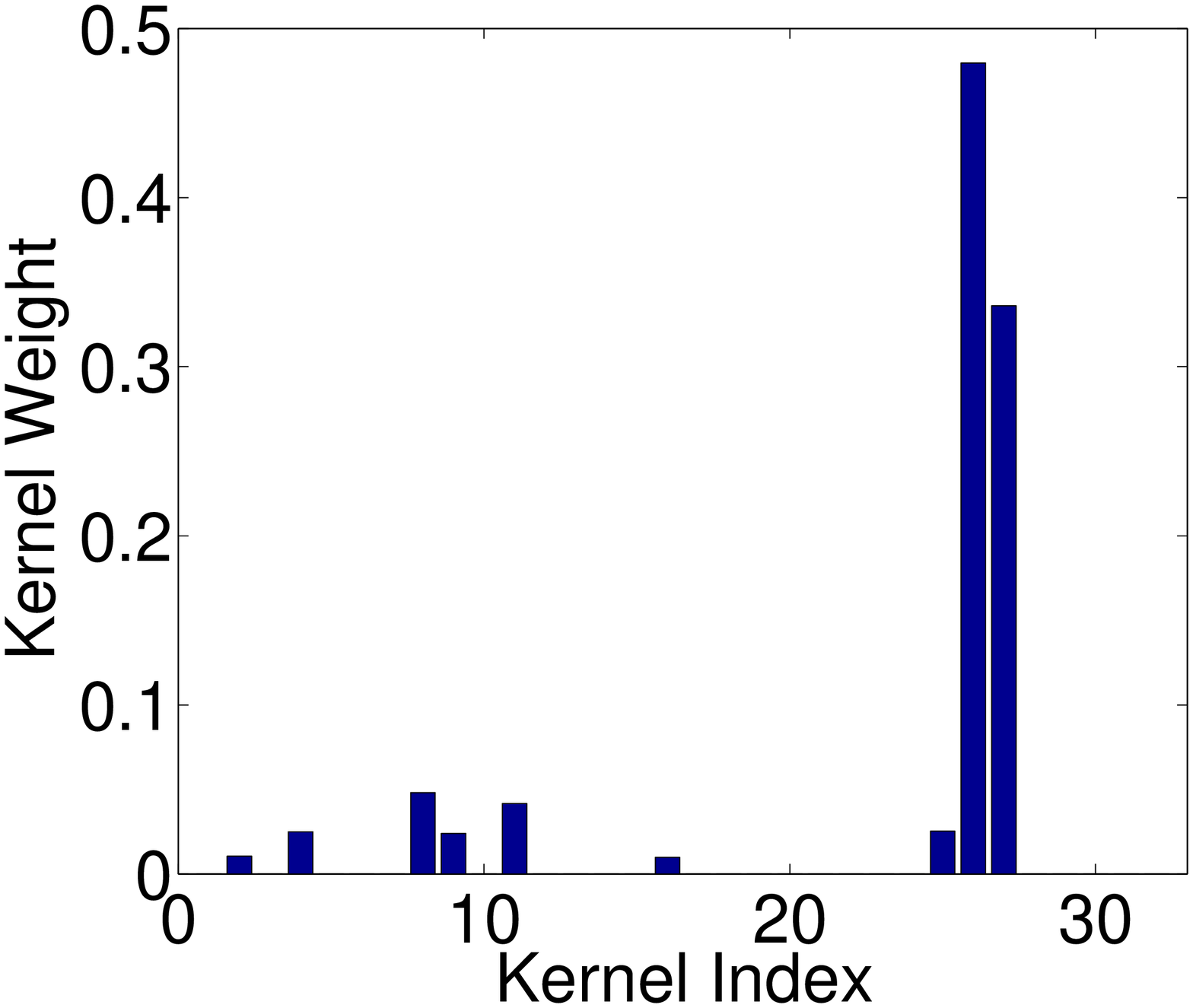} 
   \includegraphics[width=0.24\textwidth]{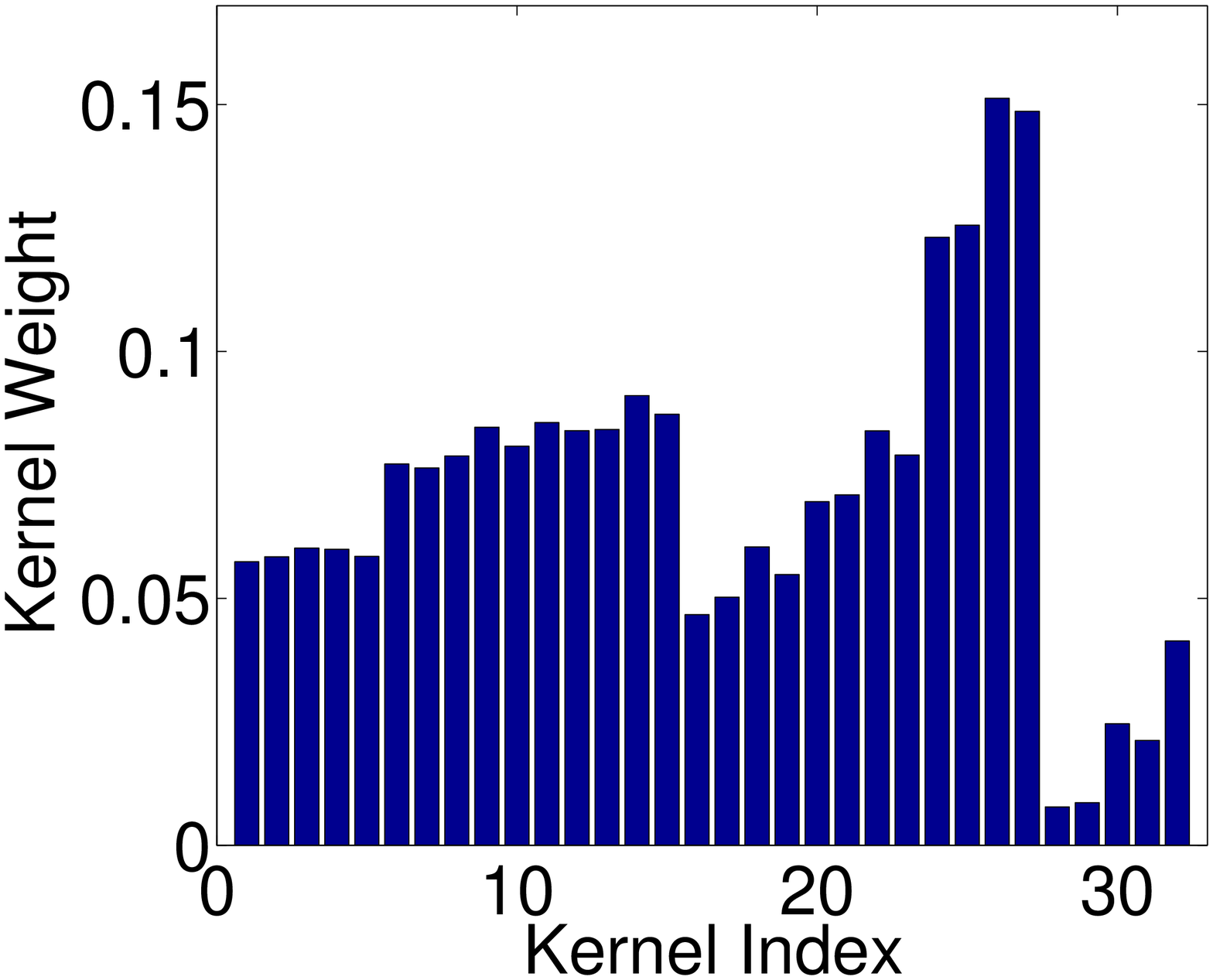}
  \caption{\small\label{fig:example_cow} Images of a typical highly ranked cow image and kernel weights from $\ell_1$-MKL (left) and $\ell_{1.333}$-MKL (right).}
  \end{center}
\end{figure}

Next, we turn to the \emph{cow} class, for which we have seen above that $\ell_{1.333}$-norm MKL outperforms all other methods clearly. Fig.~\ref{fig:example_cow} shows a typical high-ranked image of that class and also the corresponding kernel weights as output by $\ell_1$-norm (\textsc{Left}) and $\ell_{1.333}$-norm (\textsc{Right}) MKL, respectively.  We observe that $\ell_1$-MKL focuses on the two HoC product kernels; this is justified by typical cow images having green grass in the background. This allows the HoC kernels to easily to distinguish the cow images from the indoor and vehicle classes such as \emph{car} or \emph{sofa}. However, horse and sheep images have such a green background, too. They differ in sheep usually being black-white, and horses having a brown-black color bias (in VOC data); cows have rather variable colors. Here, we observe that the rather complex yet somewhat color-based BoW-C and BoW-S features help performance---it is also those kernels that are selected by the non-sparse $\ell_{1.333}$-MKL, which is the best performing model on those classes. In contrast, the sum-kernel SVM suffers from including the five gray-channel-based features, which are hardly useful for the horse and sheep classes and mostly introduce additional noise. MKL (all variants) succeed in identifying those kernels and assign those kernels with low weights.


\section{Discussion}\label{section:disc_toy}
%%%%%%%%%%%%%%%%%%%%%%%%%%%%%%%%%%%%%%%%%%%%%%%%%%%%%%%%%%%%%%%%%%%%%%%%%%%%%%%%%%%%%%%%%%

In the previous section we presented empirical evidence that $\ell_p$-norm MKL considerably can help performance in visual image categorization tasks. We also observed that the gain is class-specific and limited for some classes when compared to the sum-kernel SVM, see again Tables \ref{tab:VOC2009traincv} and \ref{tab:VOC2009test} as well as Tables B.1, B.2 in the supplemental material. In this section, we aim to shed light on the reasons of this behavior, in particular discussing strengths of the average kernel in Section \ref{ssec:argumentproavgkernel}, trade-off effects in Section \ref{ssec:priorknowledge} and strengths of MKL in Section \ref{ssec:promkl}. Since these scenarios are based on statistical properties of kernels which can be observed in concept recognition tasks within computer vision we expect the results to be transferable to other algorithms which learn linear models over kernels such as \cite{Yan:cvpr2010,DBLP:conf/iccv/CaoLLH09}.

%-------------------------------
\subsection{One Argument For the Sum Kernel: Randomness in Feature Extraction}\label{ssec:argumentproavgkernel}
%-------------------------------

We would like to draw attention to one aspect present in BoW features, namely the amount of randomness induced by the visual word generation stage acting as noise with respect to kernel selection procedures.

\smallparagraph{Experimental setup}
We consider the following experiment, similar to the one undertaken in \cite{gehler_kernelfigureitout_cvpr2009}: we compute a BoW kernel ten times each time using the same local features, identical spatial pyramid tilings, and identical kernel functions; the only difference between subsequent repetitions of the experiment lies in the randomness involved in the generation of the codebook of visual words. Note that we use SIFT features over the gray channel that are densely sampled over a grid of step size six, 512 visual words (for computational feasibility of the clustering), and a $\chi^2$ kernel. This procedure results in ten kernels that only differ in the randomness stemming from the codebook generation. We then compare the performance of the sum-kernel SVM built from the ten kernels to the one of the best single-kernel SVM determined by cross-validation-based model selection. 

In contrast to \cite{gehler_kernelfigureitout_cvpr2009} we try \emph{two} codebook generation procedures, which differ by their intrinsic amount of randomness: first, we deploy $k$-means clustering, with random initialization of the centers and a bootstrap-like selection of the best initialization (similar to the option 'cluster' in MATLAB's $k$-means routine). Second,
we deploy \emph{extremely randomized clustering forests} (ERCF) \cite{Moosmann2008,Moosmann2006}, that are, ensembles of randomized trees---the latter procedure involves a considerably higher amount of randomization compared to $k$-means.

\smallparagraph{Results}
The results are shown in Table~\ref{tab:voc10cbs}. For both clustering procedures, we observe that the sum-kernel SVM outperforms the best single-kernel SVM. In particular, this confirms earlier findings of \cite{gehler_kernelfigureitout_cvpr2009} carried out for $k$-means-based clustering. We also observe that the difference between the sum-kernel SVM and the best single-kernel SVM is much more pronounced for ERCF-based kernels---we conclude that this stems from a higher amount of randomness is involved in the ERCF clustering method when compared to conventional $k$-means. The standard deviations of the kernels in Table~\ref{tab:voc10cbs} confirm this conclusion. For each class we computed the conditional standard deviation
\begin{equation}
\label{eq:conditional_standarddeviation}
\mathrm{std}(K \mid y_i=y_j ) + \mathrm{std}(K \mid y_i \neq y_j )
\end{equation}
averaged over all classes. The usage of a conditional variance estimator is justified because the ideal similarity in kernel target alignment (cf. equation \eqref{eq:kta_idealsimilarity}) does have a variance over the kernel as a whole however the conditional deviations in equation \eqref{eq:conditional_standarddeviation} would be zero for the ideal kernel. Similarly, the fundamental MKL optimization formula \eqref{eq:mklinformation} relies on a statistic based on the two conditional kernels used in formula \eqref{eq:conditional_standarddeviation}. Finally, ERCF clustering uses label information. Therefore averaging the class-wise conditional standard deviations over all classes is not expected to be identical to the standard deviation of the whole kernel.

We observe in Table~\ref{tab:voc10cbs} that the standard deviations are lower for the sum kernels. Comparing ERCF and k-means shows that the former not only exhibits larger absolute standard deviations but also greater differences between single-best and sum-kernel as well as larger differences in AP scores. 

We can thus postulate that the reason for the superior performance of the sum-kernel SVM stems from averaging out the randomness contained in the BoW kernels (stemming from the visual-word generation). This can be explained by the fact that averaging is a way of reducing the variance in the predictors/models \cite{Breiman1996}. We can also remark that such variance reduction effects can also be observed when averaging BoW kernels with varying color combinations or other parameters; this stems from the randomness induced by the visual word generation.

Note that in the above experimental setup each kernel uses the \emph{same} information provided via the local features. Consequently, the best we can do is \emph{averaging}---learning kernel weights in such a scenario is likely to suffer from overfitting to the noise contained in the kernels and can only decrease performance.

\begin{table}
\small
\centering
\begin{tabular}{c c c}
  \hline
  Method & Best Single Kernel & Sum Kernel \\
  \hline
  VOC09, k-Means & AP: 44.42 $\pm$ 12.82 & \textbf{45.84} $\pm$ 12.94 \\
  VOC09, k-Means & Std: \textbf{30.81} & 30.74 \\
  VOC09, ERCF & AP: 42.60 $\pm$ 12.50 & \textbf{47.49} $\pm$ 12.89 \\
  VOC09, ERCF & Std: \textbf{38.12} & 37.89 \\
  ImageCLEF, k-Means& AP: 31.09 $\pm$ 5.56 & \textbf{31.73} $\pm$ 5.57 \\
  ImageCLEF, k-Means& Std: \textbf{30.51} & 30.50 \\
  ImageCLEF, ERCF& AP: 29.91 $\pm$ 5.39 & \textbf{32.77} $\pm$ 5.93 \\
  ImageCLEF, ERCF& Std: \textbf{38.58} & 38.10 \\
\hline
\end{tabular}
\caption{\label{tab:voc10cbs}\small AP Scores and standard deviations showing amount of randomness in feature extraction:results from repeated computations of BoW Kernels with randomly initialized codebooks}
\end{table}

To further analyze this, we recall that, in the computational optimum, the information content of a kernel is measured by $\ell_p$-norm MKL via the following quantity, as proved in \cite{KloBreSonZie11}:
\begin{align}\label{eq:mklinformation_aux}
   \beta ~ & \propto ~ \|w\|_2^{\frac{2}{p+1}} ~ =~ \Bigg( \sum_{i,j} \alpha_i y_i K_{ij}  \alpha_j y_j \Bigg)^{\frac{2}{p+1}} \thinspace.
\end{align}
In this paper we deliver a novel interpretation of the above quantity; to this end, we decompose the right-hand term into two terms as follows:
$$ \sum_{i,j} \alpha_i y_i K_{ij}  \alpha_j y_j ~=~ \sum_{i,j \mid y_i=y_j} \alpha_i K_{ij}  \alpha_j -  \sum_{i,j \mid y_i \neq y_j} \alpha_i K_{ij}  \alpha_j\thinspace.$$
The above term can be interpreted as a difference of the support-vector-weighted sub-kernel restricted to consistent labels \emph{and} the support-vector-weighted sub-kernel over the opposing labels. Equation~\ref{eq:mklinformation_aux} thus can be rewritten as
\begin{align}\label{eq:mklinformation}
  \beta ~ \propto ~ \Bigg( \sum_{i,j \mid y_i=y_j} \alpha_i K_{ij}  \alpha_j -  \sum_{i,j \mid y_i \neq y_j} \alpha_i K_{ij}  \alpha_j\Bigg)^{\frac{2}{p+1}} .
\end{align}
Thus, we observe that random influences in the features combined with overfitting support vectors can suggest a falsely high information content in this measure for \emph{some} kernels. SVMs do overfit on BoW features. Using the scores attained on the training data subset we can observe that many classes are deceptive-perfectly predicted with AP scores fairly above 0.9. At this point, non-sparse $\ell_{p>1}$-norm MKL offers a parameter $p$ for regularizing the kernel weights---thus hardening the algorithm to become robust against random noise, yet permitting to use some degree of information given by Equation~\eqref{eq:mklinformation}.

\cite{gehler_kernelfigureitout_cvpr2009} reported in accordance to our idea about overfitting of SVMs that $\ell_2$-MKL and $\ell_1$-MKL show no gain in such a scenario while $\ell_1$-MKL even reduces performance for some datasets. This result is not surprising as the overly sparse $\ell_1$-MKL has a stronger tendency to overfit to the randomness contained in the kernels / feature generation. The observed amount of randomness in the state-of-the-art BoW features could be an explanation why the sum-kernel SVM has shown to be a quite hard-to-beat competitor for semantic concept classification and ranking problems.

%-------------------------------
\subsection{MKL and Prior Knowledge}\label{ssec:priorknowledge}
%-------------------------------

For solving a learning problem, there is nothing more valuable than \emph{prior knowledge}. Our empirical findings on the VOC2009 and ImageCLEF09 data sets suggested that our experimental setup was actually biased towards the sum-kernel SVM via usage of prior knowledge when choosing the set of kernels / image features. We deployed kernels based on four features types: BoW-S, BoW-C, HoC and HoG. However, the \emph{number} of kernels taken from each feature type is not equal. Based on our experience with the VOC and ImageCLEF challenges we used a higher fraction of BoW kernels and less kernels of other types such as histograms of colors or gradients because we already knew that BoW kernels have superior performance.

To investigate to what extend our choice of kernels introduces a bias towards the sum-kernel SVM, we also performed another experiment, where we deployed a higher fraction of weaker kernels for VOC2009. The difference to our previous experiments lies in that we summarized the 15 BOW-S kernels in 5 product kernels reducing the number of kernels from 32 to 22. The results are given in Table~\ref{tab:VOC2009moreweakkernels}; when compared to the results of the original 32-kernel experiment (shown in Table~\ref{tab:VOC2009traincv}), we observe that the AP scores are in average about 4 points smaller. This can be attributed to the fraction of weak kernels being higher as in the original experiment; consequently, the gain from using ($\ell_{1.333}$-norm) MKL compared to the sum-kernel SVM is now more pronounced: over 2 AP points---again, this can be explained by the higher fraction of weak (i.e., noisy) kernels in the working set (this effect is also confirmed in the toy experiment carried out in supplemental material: there, we see that MKL becomes more beneficial when the number of noisy kernels is increased).

In summary, this experiment should remind us that semantic classification setups  use a substantial amount of prior knowledge. Prior knowledge implies a \emph{pre-selection} of highly effective kernels---a carefully chosen set of strong kernels constitutes a bias towards the sum kernel. Clearly, pre-selection of strong kernels reduces the need for learning kernel weights; however, in settings where prior knowledge is sparse, statistical (or even adaptive, adversarial) noise is inherently contained in the feature extraction---thus, beneficial effects of MKL are expected to be more pronounced in such a scenario.

\begin{table}
\small
\centering
 \begin{tabular}{ c c c } 
 \hline Class / $\ell_p$-norm  &  $1.333$  &  $\infty$  \\ 
 \hline 
 Aeroplane   & \textbf{77.82} $\pm$ 7.701  & 76.28 $\pm$ 8.168  \\ 
  Bicycle   & \textbf{50.75} $\pm$ 11.06  & 46.39 $\pm$ 12.37  \\ 
  Bird   & \textbf{57.7} $\pm$ 8.451  & 55.09 $\pm$ 8.224  \\ 
  Boat   & \textbf{62.8} $\pm$ 13.29  & 60.9 $\pm$ 14.01  \\ 
  Bottle   & \textbf{26.14} $\pm$ 9.274  & 25.05 $\pm$ 9.213  \\ 
  Bus   & \textbf{68.15} $\pm$ 22.55  & 67.24 $\pm$ 22.8  \\ 
  Car   & \textbf{51.72} $\pm$ 8.822  & 49.51 $\pm$ 9.447  \\ 
  Cat   & \textbf{56.69} $\pm$ 9.103  & 55.55 $\pm$ 9.317  \\ 
  Chair   & \textbf{51.67} $\pm$ 12.24  & 49.85 $\pm$ 12  \\ 
  Cow   & \textbf{25.33} $\pm$ 13.8  & 22.22 $\pm$ 12.41  \\ 
  Diningtable   & \textbf{45.91} $\pm$ 19.63  & 42.96 $\pm$ 20.17  \\ 
  Dog   & \textbf{41.22} $\pm$ 10.14  & 39.04 $\pm$ 9.565  \\ 
  Horse   & \textbf{52.45} $\pm$ 13.41  & 50.01 $\pm$ 13.88  \\ 
  Motorbike   & \textbf{54.37} $\pm$ 12.91  & 52.63 $\pm$ 12.66  \\ 
  Person   & \textbf{80.12} $\pm$ 10.13  & 79.17 $\pm$ 10.51  \\ 
  Pottedplant   & \textbf{35.69} $\pm$ 13.37  & 34.6 $\pm$ 14.09  \\ 
  Sheep   & \textbf{37.05} $\pm$ 18.04  & 34.65 $\pm$ 18.68  \\ 
  Sofa   & \textbf{41.15} $\pm$ 11.21  & 37.88 $\pm$ 11.11  \\ 
  Train   & \textbf{70.03} $\pm$ 15.67  & 67.87 $\pm$ 16.37  \\ 
  Tvmonitor   & \textbf{59.88} $\pm$ 10.66  & 57.77 $\pm$ 10.91  \\ 
  \hline 
 Average   & \textbf{52.33} $\pm$ 12.57  & 50.23 $\pm$ 12.79  \\ 
  \hline 
 \end{tabular}
\caption{\label{tab:VOC2009moreweakkernels}\small MKL versus Prior Knowledge: AP Scores with a smaller fraction of well scoring kernels}
\end{table}

%-------------------------------
\subsection{One Argument for Learning the Multiple Kernel Weights: Varying Informative Subsets of Data}\label{ssec:promkl}
%-------------------------------

In the previous sections, we presented evidence for why the sum-kernel SVM is considered to be a strong learner in visual image categorization. Nevertheless, in our experiments we observed gains in accuracy by using MKL for many concepts. In this section, we investigate causes for this performance gain. 

Intuitively speaking, one can claim that the kernel non-uniformly contain varying amounts of information content. We investigate more specifically what information content this is and why it differs over the kernels. Our main hypothesis is that common kernels in visual concept classification are informative with respect to varying subsets of the data. This stems from features being frequently computed from many combinations of color channels. We can imagine that blue color present in the upper third of an image can be crucial for prediction of photos having clear sky, while other photos showing a sundown or a smoggy sky tend to contain white or yellow colors; this means that a particular kernel / feature group can be crucial for some images, while it may be almost useless---or even counterproductive---for others.

However, the information content is accessed by MKL via the quantity given by Eq.~\eqref{eq:mklinformation}; the latter is a \emph{global} information measure, which is computed over the support vectors (which in turn are chosen over the \emph{whole dataset}). In other words, the kernel weights are global weights that uniformly hold in all regions of the input space. Explicitly finding informative subsets of the input space on real data may not only imply a too high computational burden (note that the number of partitions of an $n$-element training set is exponentially in $n$) but also is very likely to lead to overfitting. 

To understand the implications of the above to computer vision, we performed the following toy experiment. We generated a fraction of $p_+=0.25$ of positively labeled and $p_-=0.75$ of negatively labeled $6m$-dimensional training examples (motivated by the unbalancedness of training sets usually encountered in computer vision) in the following way: the features were divided in $k$ feature groups each consisting of six features. For each feature group, we split the training set into an informative and an uninformative set (the size is varying over the  feature groups); thereby, the informative sets of the particular feature groups are disjoint. Subsequently, each feature group is processed by a Gaussian kernel, where the width is determined heuristically in the same way as in the real experiments shown earlier in this paper. 

Thereby, we consider two experimental setups for sampling the data, which differ in the number of employed kernels $m$ and the sizes of the informative sets. In both cases, the informative features are drawn from two sufficiently distant normal distributions (one for each class) while the uninformative features are just Gaussian noise (mixture of Gaussians).
The experimental setup of the first experiment can be summarized as follows:

\begin{center}
  %\colorbox{lightgray}{
  %\begin{minipage}{0.465\textwidth}
\textbf{Experimental Settings for Experiment 1 (3 kernels):}
\begin{equation}
 n_{k=1,2,3}=(300,300,500) ,\ p_+\mathrel{\mathop:}=P(y=+1)=0.25
\end{equation}
The features for the informative subset are drawn according to
\begin{equation}f^{(k)}_i \sim \begin{cases}N(0.0,\sigma_k) & \mbox{ if } y_i=-1 \\ N(0.4,\sigma_k) & \mbox{ if } y_i=+1 \end{cases}\end{equation}
\begin{equation}\sigma_k=\begin{cases} 0.3 & \mbox{ if }k=1,2\\ 0.4 & \mbox{ if }k=3 \end{cases}\end{equation}
The features for the uninformative subset are drawn according to
\begin{equation}f^{(k)} \sim  (1-p_+) N(0.0,0.5) + p_+N(0.4,0.5).\end{equation} 
  %\end{minipage}
  %}
\end{center} 
%\bigskip
%\vspace{-2.9cm}
For Experiment 1 the three kernels had disjoint informative subsets of sizes $n_{k=1,2,3}=(300,300,500)$. We used $1100$ data points for training and the same amount for testing. We repeated this experiment $500$ times with different random draws of the data.

Note that the features used for the uninformative subsets are drawn as a mixture of the Gaussians used for the informative subset, but with a higher variance, though. The increased variance encodes the assumption that the feature extraction produces unreliable results on the uninformative data subset. None of these kernels are pure noise or irrelevant. Each kernel is the best one for its own informative subset of data points. 

We now turn to the  experimental setup of the second experiment:

\begin{center}
  %\colorbox{lightgray}{
%  \begin{minipage}{0.465\textwidth}
\textbf{Experimental Settings for Experiment 2 (5 kernels):}
\begin{align*}
 n_{k=1,2,3,4,5}=&\thinspace (300,300,500,200,500) ,\\
 \ p_+\mathrel{\mathop:}=&\thinspace P(y=+1)=0.25
\end{align*}
The features for the informative subset are drawn according to
\begin{equation}f^{(k)}_i \sim \begin{cases}N(0.0,\sigma_k) & \mbox{ if } y_i=-1 \\ N(m_k,\sigma_k) & \mbox{ if } y_i=+1 \end{cases}\end{equation}
\begin{equation}m_k=\begin{cases} 0.4 & \mbox{ if }k=1,2,3\\ 0.2 & \mbox{ if }k=4,5 \end{cases}\end{equation}
\begin{equation}\sigma_k=\begin{cases} 0.3 & \mbox{ if }k=1,2\\ 0.4 & \mbox{ if }k=3,4,5 \end{cases}\end{equation}
The features for the uninformative subset are drawn according to \begin{equation}f^{(k)} \sim  (1-p_+) N(0.0,0.5) + p_+N(m_k,0.5)
\end{equation}
%  \end{minipage}
  %}
\end{center} 

%\bigskip\noindent
As for the real experiments, we normalized the kernels to having standard deviation 1 in Hilbert space and optimized the regularization constant by grid search in $C\in\{10^{i}\thinspace|\thinspace i=-2,-1.5,\ldots,2\}$.

Table~\ref{tab:mkltoyexp2} shows the results. The null hypothesis of equal means is rejected by a t-test with a p-value of $0.000266$ and $0.0000047$, respectively, for Experiment 1 and 2, which is highly significant.

The design of the Experiment 1 is no exceptional lucky case: we observed similar results when using more kernels; the performance gaps then even increased. Experiment 2 is a more complex version of  Experiment 1 using  using five kernels instead of just three. Again, the  informative subsets are disjoint, but this time of sizes $300$, $300$, $500$, $200$, and $500$; the the Gaussians are centered at $0.4$, $0.4$, $0.4$, $0.2$, and $0.2$, respectively, for the positive class; and the variance is taken as $\sigma_k=(0.3,0.3,0.4,0.4,0.4)$. Compared to Experiment 1, this results in even bigger performance gaps between the sum-kernel SVM and the non-sparse $\ell_{1.0625}$-MKL. One can imagine to create learning scenarios with more and more kernels in the above way, thus increasing the performance gaps---since we aim at a relative comparison, this, however, would not further contribute to validating or rejecting our hypothesis.

\begin{table}[tbp]
\small\centering
\begin{tabular}{c c c c }
\hline
Experiment & $\ell_{\infty}$-SVM & $\ell_{1.0625}$-MKL & t-test p-value \\ \hline
 1& 68.72 $\pm$ 3.27 & 69.49 $\pm$ 3.17 & 0.000266\\ %callgendata1,evalscript: data3
2 & 55.07 $\pm$ 2.86 & 56.39 $\pm$ 2.84 & $4.7\cdot10^{-6}$\\ %callgendata2,evalscript: data5
\hline
\end{tabular}
\caption{\label{tab:mkltoyexp2}\small Varying Informative Subsets of Data: AP Scores in Toy experiment using Kernels with disjoint informative subsets of Data}
\end{table}

Furthermore, we also investigate the single-kernel performance of each kernel:  we observed the best single-kernel SVM (which attained AP scores of  $43.60$, $43.40$, and $58.90$ for Experiment 1) being inferior to both MKL (regardless of the employed norm parameter $p$) and the sum-kernel SVM. The differences were significant with  fairly small p-values (for example, for $\ell_{1.25}$-MKL the p-value was about $0.02$).

We emphasize that we did not design the example in order to achieve a maximal performance gap between the non sparse MKL and its competitors. For such an example, see the toy experiment of \cite{KloBreSonZie11}, which is replicated in the supplemental material including additional analysis. Our focus here was to confirm our hypothesis that kernels in semantic concept classification are based on varying subsets of the data---although MKL computes global weights, it emphasizes on kernels that are relevant on the largest informative set and thus approximates the infeasible combinatorial problem of computing an optimal partition/grid of the space into regions which underlie identical optimal weights. Though, in practice, we expect the situation to be more complicated as informative subsets may overlap between kernels.
 
Nevertheless, our hypothesis also opens the way to new directions for learning of kernel weights, namely restricted to subsets of data chosen according to a meaningful principle. Finding such principles is one the future goals of MKL---we sketched one possibility: locality in feature space. A first starting point may be the work of \cite{gonen10icpr,DBLP:conf/iccv/YangLTDG09} on localized MKL.

%%%%%%%%%%%%%%%%%%%%%%%%%%%%%%%%%%%%%%%%%%%%%%%%%%%%%%%%%%%%%%%%%%%%%%%%%%%%%%%%%%%%%%%%%%
\section{Conclusions}\label{conclusion}
%%%%%%%%%%%%%%%%%%%%%%%%%%%%%%%%%%%%%%%%%%%%%%%%%%%%%%%%%%%%%%%%%%%%%%%%%%%%%%%%%%%%%%%%%%

When measuring data with different measuring devices, it is always a challenge to combine the respective devices' uncertainties in order to fuse all available sensor information optimally. In this paper, we revisited this important topic and discussed machine learning approaches to adaptively combine different image descriptors in a systematic and theoretically well founded manner. While MKL approaches in principle solve this problem it has been observed that the standard $\ell_1$-norm based MKL often cannot outperform SVMs that use an average of a large number of kernels. One hypothesis why this seemingly unintuitive result may occur is that the sparsity prior may not be appropriate in many real world problems---especially, when prior knowledge is already at hand. We tested whether this hypothesis holds true for computer vision and applied the recently developed non-sparse $\ell_p$ MKL algorithms to object classification tasks. The $\ell_p$-norm constitutes a slightly less severe method of sparsification. By choosing $p$ as a hyperparameter, which controls the degree of non-sparsity and regularization, from a set of candidate values with the help of a validation data, we showed that $\ell_p$-MKL significantly improves SVMs with averaged kernels and the standard sparse $\ell_1$ MKL.

Future work will study localized MKL and methods to include hierarchically structured information into MKL, e.g. knowledge from taxonomies, semantic information or spatial priors. Another interesting direction is MKL-KDA \cite{Yan:icdm09,Yan:cvpr2010}. The difference to the method studied in the present paper lies in the base optimization criterion: KDA \cite{MikRaeWesSchMue99} leads to non-sparse solutions in $\valpha$ while ours leads to sparse ones (i.e., a low number of support vectors). While on the computational side the latter is expected to be advantageous, the first one might lead to more accurate solutions. We expect the regularization over kernel weights (i.e., the choice of the norm parameter $p$) having similar effects for MKL-KDA like for MKL-SVM. Future studies will expand on that topic.
\footnote{First experiments on ImageCLEF2010 show for sum kernel SRKDA \cite{CHH07i} a result of 39.29 AP points which is slighlty better than the sum kernel results for the SVM (39.11 AP) but worse than MKL-SVM.}

%%%%%%%%%%%%%%%%%%%%%%%%%%%%%%%%%%%%%%%%%%%%%%%%%%%%%%%%%%%%%%%%%%%%%%%%%%%%%%%%%%%%%%%%%%
\section*{Acknowledgments}
%%%%%%%%%%%%%%%%%%%%%%%%%%%%%%%%%%%%%%%%%%%%%%%%%%%%%%%%%%%%%%%%%%%%%%%%%%%%%%%%%%%%%%%%%%

This work was supported in part by the Federal Ministry of Economics and Technology of Germany (BMWi) under the project THESEUS (FKZ 01MQ07018),  by Federal Ministry of Education and Research (BMBF) under the project REMIND (FKZ 01-IS07007A), by the Deutsche Forschungsgemeinschaft (DFG), and by the FP7-ICT program of the European Community, under the PASCAL2  Network of Excellence (ICT-216886). Marius Kloft acknowledges a scholarship by the German Academic Exchange Service (DAAD).

\bibliographystyle{plos2009}
{\small
\bibliography{MachineLearning,ImageAnalysis}
}

\end{document}